\documentclass[10pt,journal,compsoc]{IEEEtran}

\usepackage{pgfplots}
\usepackage{graphicx}
\usepackage{multicol,multirow}
\usepackage{booktabs}
\usepackage{color}
\usepackage{amsmath}
\usepackage{xspace}
\usepackage{subcaption}
\usepackage{caption}
\usepackage{enumitem}
\usepackage{amsfonts}
\usepackage[normalem]{ulem}

\usepackage{sidecap}

\newcommand*{\eg}{e.g.\@\xspace}
\newcommand*{\ie}{i.e.\@\xspace}
\newcommand*{\etal}{et al.\@\xspace}
\graphicspath{{images/}}

\newcommand{\djt}[1]{\textcolor{red}{{\bf David:} #1}}

\newcommand{\yida}[1]{\textcolor{violet}{{\bf Yida:} #1}}
\newcommand{\todo}[1]{\textcolor{blue}{{\bf TODO:} #1}}

\newcommand{\figref}[1]{Fig.~\ref{#1}}
\newcommand{\secref}[1]{Sec.~\ref{#1}}
\newcommand{\tabref}[1]{Table~\ref{#1}}

\ifCLASSOPTIONcompsoc
  \usepackage[nocompress]{cite}
\else
  \usepackage{cite}
\fi
\ifCLASSINFOpdf
\else
\fi

\begin{document}
%
\title{Self-supervised Latent Space Optimization \\ with Nebula Variational Coding}
%
%
%
%

\author{Yida~Wang,
        David~Joseph~Tan,
        Nassir~Navab,
        and~Federico~Tombari 
}

\IEEEtitleabstractindextext{
\begin{abstract}
Deep learning approaches process data in a layer-by-layer way with intermediate (or latent) features. 
We aim at designing a general solution to optimize the latent manifolds to improve the performance on classification, segmentation, completion and/or reconstruction through probabilistic models.
This paper proposes a variational inference model which leads to a clustered embedding.
We introduce additional variables in the latent space, called \emph{nebula anchors}, that guide the latent variables to form clusters during training.
To prevent the anchors from clustering among themselves, we employ the variational constraint that enforces the latent features within an anchor to form a Gaussian distribution, resulting in a generative model we refer as Nebula Variational Coding (NVC).
Since each latent feature can be labeled with the closest anchor, we also propose to apply metric learning in a self-supervised way to make the separation between clusters more explicit. 
As a consequence, 
the latent variables of our variational coder form clusters which adapt to the generated semantic of the training data, \eg the categorical labels of each sample. 
We demonstrate experimentally that it can be used within different architectures designed to solve different problems including text sequence, images, 3D point clouds and volumetric data, validating the advantage of our proposed method. 
\end{abstract}

}

\maketitle

\IEEEdisplaynontitleabstractindextext

%
\IEEEpeerreviewmaketitle

\IEEEraisesectionheading{\section{Introduction}\label{sec:introduction}}

\IEEEPARstart{I}{n}
machine learning, we aim at learning relationships of different kinds between the input and the output signals.
For instance, solutions in 
language translation~\cite{lewis2019bart, lample2019cross} and 
3D understanding~\cite{dai2017shape, yuan2018pcn, wang2019forknet} 
aim at completely transforming the input data to a different form as the probabilities in classification and segmentation.
They rely on compressing the input to its latent representations while identifying its discriminative information.
On the other hand, solutions in image processing~\cite{yuan2020efficient, bao2020real, abrevaya2020cross, azad2019bi} and 
3D understanding~\cite{yang2018dense, liu2020morphing, grnet_xie, Wen_2020_CVPR, wang2020deep, abrevaya2020cross} 
aim at preserving the information from the input data as part of the output, formulating additional skip connection~\cite{azad2019bi, abrevaya2020cross, yang2018dense, grnet_xie} and fusion~\cite{liu2020morphing} between them.


Although there are traditional methods that can reduce the dimensionality of the input in an unsupervised way, such as 
principal component analysis (PCA)~\cite{bell1995information} and 
independent component analysis (ICA)~\cite{hyvarinen1999fixed}, as well as in a supervised way, e.g. linear discriminant analysis (LDA)~\cite{swets1996using}, 
they can hardly be applied when modeling complex data. 

Moving towards deep learning, such compression can also be modeled with an encoder-decoder architecture. 
Auto-encoders (AE)~\cite{vincent2010stacked, makhzani2015adversarial} are among the simplest, yet effective, architectures able to extract the latent features in an unsupervised fashion. For these reasons, they are commonly employed in a variety of tasks such as image denoising~\cite{bengio2013generalized, im2017denoising} and retrieval~\cite{yun2021pairwise}.
Moreover, more complicated tasks such as pose estimation~\cite{GANeratedHands_CVPR2018, zimmermann2017learning, gao2019variational} or 3D completion~\cite{wang2019forknet, yang2018dense} utilize a more generalized encoder-decoder architecture.

\begin{figure*}[t]
\centering
\includegraphics[width=\linewidth]{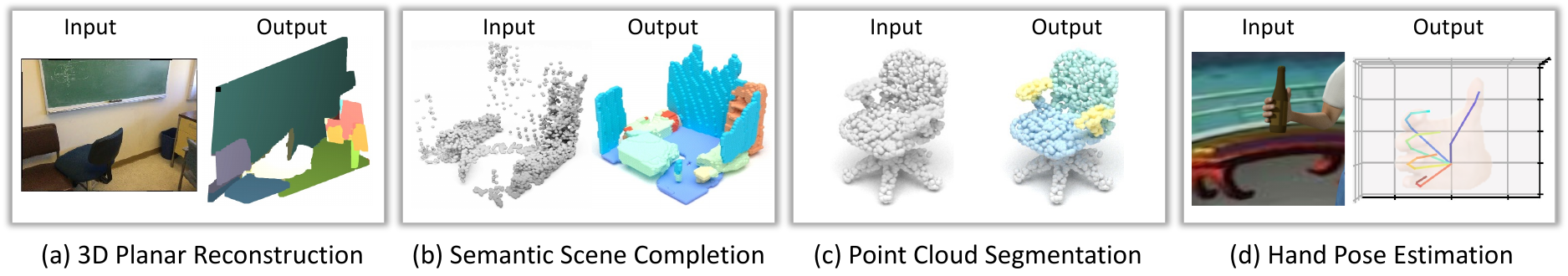}
\caption{Example applications of the proposed nebula variational coder (NVC) on different architectures that are designed to solve problems in
(a)~planar reconstruction, 
(b)~semantic scene completion,
(c)~point cloud segmentation
and
(d)~hand pose estimation.
Note that (a) and (b) take real images captured from real scenes as input while (c) and (d) take synthetic images rendered from realistic scenes.
}
\label{fig:teaser}
\end{figure*}



Looking at the bigger picture, there are three ways to improve the architecture, namely, encoder design~\cite{patacchiola2020autoencoders}, decoder design~\cite{chen2019bae} and latent feature optimization~\cite{patacchiola2020autoencoders}.
In this work, we focus on proposing a novel latent feature optimization approach.

One of the popular approaches to optimize the latent feature is through the variational inference~\cite{rolinek2019variational, Sohn2015cgm}. 
It matches the likelihood of the encoder with the true posterior of the decoder by setting a distribution constraint on the latent features such as Gaussian or Bernoulli.
Since they impose a distribution in the latent space, it becomes feasible to generate reasonable output, with a trained decoder, directly from random sampling of the latent feature, transforming the deep architecture to a generative model such as variational auto-encoder (VAE)~\cite{rolinek2019variational, Sohn2015cgm}.


However, there are two main problems in assuming a simple distribution like the Gaussian in VAEs. 
First, the latent space often includes unused regions which are accountable for generating samples that do not belong to the real distribution. 
Specifically for VAEs, the Gaussian assumption with the fixed expected density does not always match the likelihood of the encoder with the true posterior of the decoder, leading, \eg, to blurry reconstructed contours in perceptual VAE~\cite{dosovitskiy2016generating}. Such problem is mostly influenced by the decoder.
With respect to the encoder, another problem is the uncertainty described in~\cite{walker2016uncertain}, where the latent variables are not expected to be continuous in some situations. 


To solve these issues, conditional information~\cite{Sohn2015cgm} and additional parameters~\cite{wang2019forknet, yang2018dense} out of the encoder-decoder inference model are used to improve their results.
Nevertheless, these solutions carry their own disadvantages. In the former, the conditional information is still required during inference; while, in the latter, the additional architecture makes training less efficient.


We propose a new variational inference model which can be applied to different architectures, and used in a variety of supervised and unsupervised training including classification, regression and recurrent prediction.
Our approach relies on the idea of having additional latent variables used during training, called \emph{nebula anchors}, that are explicitly embedded in training the model so to help the formation of hyper-clusters in the latent space.
One of its main advantages is the ability to exploit the semantic meaning such as the categorical information in the latent space without any human supervision.
As an example on the MNIST dataset~\cite{lecun1998mnist}, we illustrate that our generative model automatically forms 10 clusters, each representing a different digit (see \figref{fig:exp_mnist_orders} (b)).

The limitation of our anchor-based solution is the need to pre-define one key hyper-parameter which is the number of anchors. To ease this, 
we also introduce a self-supervised metric learning derived from the Siamese~\cite{lu2014neighborhood} and Triplet~\cite{wohlhart15, kumar2016learning, movshovitz2017no} losses, tailored for the nebula anchors.
We apply the metric learning efficiently using the labels produced by the nebula anchors based on the nearest neighbour classifier to separate the clusters.
Notably, unlike other generative models such as CVAE~\cite{Sohn2015cgm} that use labels as conditional variables to improve inference, our method exploits the additional information only during training, which implies that no additional information other than the input samples is required during inference.

By design, our nebula anchors can be applied on the latent space of not just the VAEs, but also other deep networks such as Convolutional Neural Networks (CNN)~\cite{Krizhevsky2012, szegedy2015going}, Recurrent Neural Networks (RNNs)~\cite{sutskever2014sequence,britz2017massive} as well as adversarial generative models such as GANs~\cite{NIPS2014_5423,maaloe2016auxiliary,wang2019forknet}.
Motivated by the range of applicability, we evaluate the proposed approach on various datasets that include image reconstruction on 
language translation on WMT16~\cite{bojar2016findings}, 
image reconstruction on MNIST~\cite{lecun1998mnist}, 
3D volumetric completion on objects from ShapeNet~\cite{yang2018dense}, 
3D point cloud segmentation on PointNet~\cite{qi2017pointnet},
3D hand pose estimation on HOP~\cite{gao2019variational} and Stereo~\cite{zhang20163d},
3D planar reconstruction on NYUv2~\cite{yu2019single} and ScanNet\cite{dai2017scannet},
and
3D semantic completion on ScanNet~\cite{dai2017scannet}.
Some of these results are shown in \figref{fig:teaser}. 
Strikingly, even if our experiments evaluate on varying datasets with distinct objectives and dimensionalities such as sentences, images and 3D point clouds, we demonstrate that our model is beneficial in the encoder-decoder architectures.

\section{Related Works}
\label{sec:related}

An encoder-decoder architecture is designed to extract the latent features from the input which are then mapped to the output.
%
Auto-encoders (AEs)~\cite{choi2019variable, zhang2020uc} are among the simplest encoder-decoder architectures. 
Their features are used to solve a large variety of problems in image compression~\cite{choi2019variable}, video compression~\cite{habibian2019video}, anomaly detection~\cite{gong2019memorizing}, saliency detection~\cite{zhang2020uc} and 3D segmentation~\cite{chen2019bae}. 
By making the parametric model deeper such as stacked convolutional AE~\cite{masci2011stacked}, wider such as BAE-Net~\cite{chen2019bae} or nested such as AE$^2$-Nets~\cite{zhang2019ae2}, they improve ability of AEs to fit the training data. 



While AEs are trained in an unsupervised way, there are also architectures that use supervised information to help train the latent features. Those works show obvious advantages against traditional approaches~\cite{chan2015pcanet}. 
%
%
%
Tackling a more complex problem like reconstructing 3D structures from a single 2D image, Deep Supervision~\cite{li2017deep} utilizes the additional skeleton labels to optimize the output of the hidden layers without using the variational methods to estimate the occluded area in the 2D image. 
Recently, there are more works in self-supervised learning for image classification~\cite{gidaris2019boosting}, point cloud retrieval~\cite{wang2021self} and 3D Axon Segmentation~\cite{klinghoffer2020self}. 

Ranging from a simple auto-encoder to a more complicated encoder-decoder architecture, the proposed nebula anchors can be integrated in all of them. 
%
%
Particularly, this work proposes a general optimization approach to improve the latent space in an encoder-decoder architecture which can be applied to different problems. 
It is noteworthy to mention that our experiments in \secref{sec:experiments} evaluates on language translation, image reconstruction, and 3D reconstruction and pose estimation, where
we demonstrate the superiority of incorporating the nebula anchors in the existing architectures.


The related work on the latent space optimization can be categorize into four main fields, namely, variational embedding, clustering, metric learning and adversarial learning. 
The following sections discuss them in more detail.
Compared to these fields, we propose a variational approach that forms clusters in the latent space while incorporating metric learning as an optional optimization approach to further improve the performance.

\subsection{Variational embedding}
\label{sec:variational_inference}

Given an encoder-decoder architecture, the variational inference use the encoder to approximate the posterior of the latent features that are conditioned on the expected network output. This is governed by a simple distribution of the data in latent space. 
These methods are derived from VAE, where \cite{rolinek2019variational} proposes an assumption that the latent feature of VAE behaves like PCA components.
This, in effect, makes the convergence in training more efficient.

To solve the problem caused by the simplicity of the latent space of VAE, categorical labels are integrated in conditional VAE (CVAE),
which is used in generation~\cite{Sohn2015cgm} and prediction~\cite{walker2016uncertain}.
With a better performance than the standard VAE, they prove that such information holds the potential to improve the generative models. 

Using variational inference in clustering, GMVAE~\cite{dilokthanakul2016deep} and its graph embedding version~\cite{yang2019deep} demonstrate that multi-modal Gaussian can reveal categorical information better than VAE.
Another notable method is from VQ-VAE~\cite{van2017neural} which aims at solving the ``posterior collapse" by discretizing the trained features into a table.
Moreover, InfoVAE~\cite{zhao2019infovae} improves the evidence lower bound (ELBO) objective since they observed that the ELBO favors in optimizing the distribution over the inference.

Going beyond the assumption of simple distributions in the latent space, Auxiliary Deep Generative Models~\cite{maaloe2016auxiliary} adds auxiliary latent variables. But the disadvantage of such work is that the auxiliary latent variables are necessary not just during training but also at inference time.

Apart from dealing with the visual data, VAE can also be used for natural language processing (NLP) such as machine translation. 
VAE-LSTM~\cite{bowman2016generating} composes both the encoder and the decoder with LSTM operations while VAE-CNN~\cite{yang2017improved} uses LSTM in the encoder and the dilated CNN to form the decoder. 
In these methods, to solve the latent variable collapse problem~\cite{bowman2016generating, li2019stable} of VAEs, HR-VAE~\cite{li2019stable} imposes regularization for all the hidden states of the LSTM encoder.

\subsection{Clustering the latent features}
\label{sec:clustering}

The main technique in the proposed method is the formation of clusters in the latent space through our nebula anchors. 
However, there are many related works focusing on building clusters in the latent space. 

As a simple deep clustering approach, DEC~\cite{xie2016unsupervised} is among the first few works that clusters the latent space in AEs. 
Aiming at including the discrete class labels into the latent space, $K$-means~\cite{Tang2015kernelKmeans} is another method to cluster the feature~\cite{Norouzi_2013_CVPR, Tang2015kernelKmeans}. For instance, DCN~\cite{yang2017towards} uses the $K$-means clusters to learn a compact feature within an AE architecture.

A notable work is from 
Gumbel-softmax~\cite{Jang2016CategoricalRW} where they train discrete latent variables by utilizing additional labels in a way similar to our work. 
But, in contrast, 
\cite{Jang2016CategoricalRW} interpolates between the discrete and continuous distributions while our loss function directly operate on the distance metric between features and clusters.
Moreover, PrototypicalNet~\cite{snell2017prototypical} builds an embedding table for the latent features which are then used to represent a feature by applying the softmax activations over distances of the feature to each vector in the table.
Moreover, the hierarchical clustering such as HG~\cite{joe10500845} can exploit hierarchically organized auxiliary labels to learn the clusters in the embedding space; while, 
IMSAT~\cite{chang2017deep} and IIC~\cite{ji2019invariant} learn the latent clusters by maximizing the sample-wise categorical information. 

In some cases, more than one labels are given for a single sample. For instance, the object in an image is labeled with the illumination and its styles. 
Targeted on this situation, the latent space are then disentangled so that the latent clusters reveal specific attributes such as in CDD~\cite{gonzalez2018image} and Y-AE~\cite{patacchiola2020autoencoders}.
Among all the disentanglement approaches, two~\cite{hadad2018two, patacchiola2020autoencoders, zheng2019disentangling} or three~\cite{gonzalez2018image} attributes are commonly investigated.

\subsection{Metric learning}
\label{sec:metric_learning}

There have been several methods that apply metric learning in the latent space optimization~\cite{kumar2016learning, wang2019multi, wu2018improving}.
Assuming a supervised learning, these methods optimizes the distance among the samples so that they reflect their ground truth semantic similarity. 
They formulate the pairwise distance metrics, which include: the triplet loss and its derivatives~\cite{kumar2016learning, hoffer2015deep, sohn2016improved, oh2016deep}, the contrastive loss and its derivatives~\cite{hadsell2006dimensionality, wang2019multi}, and the Neighborhood Component Analysis and its derivatives~\cite{goldberger2004neighbourhood, movshovitz2017no, wu2018improving}.
Among all those losses, Triplet learning~\cite{wohlhart15,kumar2016learning,WangSLRWPCW14,hoffer2015deep,schroff2015facenet} is one of the typical learning strategy where the pairwise distances are further labeled as positive or negative based on the pair-wise relationships, resulting in clusters in the latent space. 

In this method, we also employ the triplet loss. But, 
differently from them, our method can exploit the auxiliary labels even if they do not have explicit labels, \ie unsupervised learning.



\subsection{Adversarial feature learning}
\label{sec:adversarial_learning}

In some tasks, the training data is different from test data which causes a data migration problem. This implies that the feature domains extracted by the same encoder are different. 
The objective then is to make both domains similar to each others so that the network trained from one type of data could be applied to the test data. Such tasks could be referred as domain adaptation~\cite{hu2020unsupervised, tang2020unsupervised, yang2020one} or domain generalization~\cite{gong2019dlow}, and could be solved by discriminative training~\cite{schonfeld2020u, chen2020reusing}.

For instance, the domain discriminator in DANN~\cite{ganin2016domain} is used to distinguish the source domain from the target domain using a binary code, while ADDA~\cite{tzeng2017adversarial} uses two different discriminators for the source feature and the target features. 
If additional category labels are available, SymNets~\cite{zhang2019domain} can further improve the domain adaptation by using the two-level domain confusion losses from the domain level to category level.
By making the latent feature extracted by the encoder in the same dimension as the input, GVB~\cite{cui2020gradually} uses discriminative training to make the latent features from the two different domains to be in the same sub-space. 




\section{Methodology}
\label{sec:generative_model}

Given the task of estimating the expected output $Y$ from the input $X$, the architecture of generative models such as VAE~\cite{kingma2013auto} and CVAE~\cite{Sohn2015cgm} constitute two parts --
the encoder $\mathcal{E}(X)$ that compresses the input $X$ into the latent variable $z$, and 
the generator or decoder $\mathcal{G}(z)$, which maps $z$ into the output $Y$. 

During training, the parameters in the architecture are optimized through the loss function where its definition depends on the problem at hand.
For instance, regression tasks such as image reconstruction~\cite{lecun1998mnist} and language translation~\cite{wu2016google} commonly define the loss function by means of the Euclidean distance
\begin{align}
	\mathcal{L}_\text{Euclidean} = 
	\| 
	\tilde{Y} - \mathcal{G}(\mathcal{E}(X))
	\|^2
	\label{eq:l_pred_regression}
\end{align}
where we differentiate the predicted output ($Y$) from the ground truth ($\tilde{Y}$).
The solutions for the completion of partially scanned 3D volumetric objects~\cite{yang2018dense} or point cloud segmentation~\cite{qi2017pointnet} predict a one-hot encoded vector that allows training by means of a binary cross entropy loss function 
\begin{align}
	\mathcal{L}_\text{Entropy} 
	= -\tilde{Y} \log (\mathcal{G}(\mathcal{E}(X))) 
	- (1-\tilde{Y}) \log (1 - \mathcal{G}(\mathcal{E}(X)))~.
	\label{eq:l_pred_one_hot}
\end{align}
Furthermore, the point cloud reconstruction of the objects~\cite{yuan2018pcn} evaluates whether the reconstructed point cloud $\mathcal{G}(\mathcal{E}(X))$ matches the given ground truth $\tilde{Y}$ through the Chamfer distance
\begin{align}
	\mathcal{L}_\text{Chamfer} 
	= \text{Chamfer}((\mathcal{G}(\mathcal{E}(X)), \tilde{Y})~.
	\label{eq:l_pred_chamfer}
\end{align}
Later in \secref{sec:experiments}, we use in our experiments the same loss functions as in \eqref{eq:l_pred_regression}, \eqref{eq:l_pred_one_hot} and \eqref{eq:l_pred_chamfer} for the respective problems.

In most works~\cite{qi2017pointnet, dai2017shape}, 
the role of the latent feature is simplified to be the output of the encoder and the input of the generator.
However, since the latent features extract the most useful information from $X$, 
we aim at capturing the contextual information from the latent feature (and indirectly from $X$) by formulating clusters in the latent space in order to easily build the relation between the latent feature and the output.

\subsection{Learning with Nebula Anchors} 
\label{sec:nvc}

Inspired by $K$-means~\cite{Tang2015kernelKmeans} where the sampled features form clusters by iteratively updating their centers, we introduce the concept of \emph{nebula anchors} to parameterize the cluster centers. 

We define the set of the $m$ nebula anchors as $\mathcal{A} = \{a_1, a_2, \ldots, a_m\}$, each represented by a vector with the same dimension $m$ as that of the latent variable. 
In order to match them, the latent variables are labeled with the ID of the nearest nebula anchor so that the variable with the same label are expected to stay close to the associated anchor, while the anchor itself moves towards regions with higher density of variables and the same label.
We can then define $\mathcal{Z}_{a_i}$ as the set of latent features that are close to $a_i$. 
It follows that $a_i$ is the cluster center of the latent features in $\mathcal{Z}_{a_i}$.

\subsubsection{Nebula loss}

Our Nebula loss is inspired by the concept of Newton's law of universal gravitation, where the force is proportional to the two masses and inversely proportional to their squared distance.
%
In this context, we interpret the mass 
\begin{align}
{M}(a_i) = 1+
    \sum_{\mathcal{E}(X) \in z_{a_i}} \left\|\mathcal{E}(X) - a_i\right\|^2 
    \label{equ:dist}
\end{align}
as the sum of distances from each feature attracted to its anchor.
Since the anchor $a_i$ is the cluster center of the latent features in $\mathcal{Z}_{a_i}$, the value of the mass is related to the distance of the features to the anchor and the number of features in the anchor. 

\begin{figure}[t]
\centering
\begin{tikzpicture}
\tikzstyle{every node}=[font=\small]
\begin{axis}[
    xlabel={$\|a_j-a_i\|$},
    ylabel={$f(\|a_j-a_i\|)$},
    xmin=0, xmax=3,
    ymin=-1, ymax=4,
    legend pos=north east,
    legend cell align={left},
    legend style={nodes={scale=0.75, transform shape}},
    ymajorgrids=true,
    grid style=dashed,
    scale=0.55
]
\addplot [
    domain=0.001:3, 
    samples=50, 
    color=red,
    ]
    {-log10(x^2)};
\addlegendentry{$-\log\|a_j-a_i\|^2$}
\addplot [
    domain=0:3, 
    samples=50, 
    color=blue,
    ]
    {1/x^2};
\addlegendentry{$\frac{1}{\|a_j-a_i\|^2}$}
\end{axis}
\end{tikzpicture}
\caption{Comparison of $-\log\|a_j-a_i\|^2$ and $\frac{1}{\|a_j-a_i\|^2}$.}
\label{fig:inverse_distance}
\end{figure}
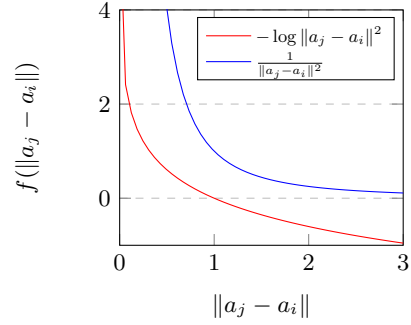

Instead of dividing by the squared distance, we use the negative logarithm of the squared distance 
\begin{align}
{D}^{-2}(a_i, a_j)=
    -\log{\|a_j - a_i\|^2}
    ~\propto~
    \frac{1}{\|a_j - a_i\|^2} ~.
    \label{equ:anchors}
\end{align}
Notably, the negative logarithm function is proportional to the inverse of the distance as shown in \figref{fig:inverse_distance}. Considering that we have the assumption of a variational latent space (see \secref{sec:variational}), the distance between a pair of anchors $\|a_j - a_i\|^2$ has a stable range roughly between 0 to 1. The main difference between the two is the capacity of the logarithmic function to enforce the optimized distance between the two anchors to be 1 instead of infinity. This, in effect, makes the optimization more stable.


Now that we have defined the mass of an anchor and inverse distance between two anchors, we build the formula for the gravitational force as 
\begin{align}
{F}(a_i, a_j) = 
	{M}(a_i) \cdot {M}(a_j) \cdot D^{-2}(a_i, a_j) ~.
	\label{eq:gravitational_force}
\end{align}
By minimizing the force, we also minimize the distance from the feature to the anchors which makes them more compact and, at the same time, maximize the distance between the anchors.
%
Therefore, to build the final loss, we sum up the gravitational forces from all pairs of anchors. The nebula loss then is 
\begin{align}
	\mathcal{L}_\text{nebula} 
	&= 
	\sum_{i=1}^{m-1}\sum_{j=1+1}^{m}{
	{F}(a_i, a_j)}  \nonumber \\
	&= 
	\sum_{i=1}^{m-1}\sum_{j=1+1}^{m}{
	{M}(a_i)  \cdot
	{{M}(a_j)} \cdot {D}^{-2}(a_i, a_j)} \nonumber \\
	&=
	\sum_{i=1}^{m-1}{
	{M}(a_i)  \cdot
	\sum_{j=1+1}^{m}{{M}(a_j)} \cdot {D}^{-2}(a_i, a_j)}
	\label{equ:nebula}
\end{align}
which ensures two criteria: (1)~the latent feature belongs to the closest anchor; and, (2)~the anchors are separated from each other.

\subsubsection{Distinction from $K$-means}

As $K$-means~\cite{Tang2015kernelKmeans} has been widely used for feature clustering \cite{Norouzi_2013_CVPR, Tang2015kernelKmeans}, we want to point out the differences from the proposed nebula loss.
If the nebula anchors are optimized by $K$-means~\cite{Tang2015kernelKmeans},
the cluster centers are then updated directly with
\begin{align}
	a_i = 
	\frac{1}{|\mathcal{Z}_{a_i}|} 
	\sum_{z_j \in \mathcal{Z}_{a_i}} z_j
\end{align}
where $|\mathcal{Z}_{a_i}|$ is the number of elements in $\mathcal{Z}_{a_i}$. 
The loss function is therefore written as 
\begin{align}
	\mathcal{L}_\text{$K$-means} = 
		\sum_{i=1}^{m} 
		\sum_{z_j \in \mathcal{Z}_{a_i}} 
		\|z_j - a_i\|^2 
		= \sum_{i=1}^{m}
		|\mathcal{Z}_{a_i}| \cdot \mathrm{Var}(\mathcal{Z}_{a_i})
\end{align}
where $\mathrm{Var}(\cdot)$ is the variance of the set.
However, since $K$-means is evaluated on all the samples to update the centers of every cluster, it could not be applied in the latent features when the network is being optimized by the Stochastic Gradient Descent (SGD)~\cite{zinkevich2010parallelized}.
One problem is that $\mathcal{L}_\text{$K$-means}$ depends on a large amount of samples, which is not accessible when we optimize the parametric model with samples in the mini-batches.

Although the Robbins-Monro stochastic approximation~\cite{robbins1951} can perform batch-wise optimization similar to SGD using
\begin{align}
	a_i^\text{new} = 
	a_i^\text{old} + 
	l_r \sum_{\mathcal{E}(X) \in z_{a_i}}(\mathcal{E}(X)-a_i^\text{old})
\end{align}
with the learning rate $l_r$,
this equation requires each subset $\mathcal{Z}_{a_i}$ to come from a fixed set of latent features with a constant total variance. 
%
Prior to convergence, the parameters in the encoder constantly change which, in effect, changes the values of the latent features across all the mini-batches during training. 

Differently from the cluster centers in $K$-means, our nebula anchors could be optimized via SGD.
To train our loss batch-wise, we make the variational assumption on the distribution of latent features and initialize the nebula anchors in a Gaussian distribution accordingly. 
Hence, this function is evaluated under unsupervised training and forces the network to create nebula anchor-driven clusters in the latent space.

\subsubsection{Variational constraint}
\label{sec:variational}

The variational inference model imposes a pre-defined distribution on the latent feature to optimize the encoder-decoder architecture.
Contrary to other methods, we implement the variation constraint through the nebula anchors.

We first adopt the variation translation model from GNMT~\cite{wu2016google} where the expected $Y$ is different from input $X$.
This implies that we can denote a given problem through the probabilistic model $P(Y|X)$.

Given the encoder $\mathcal{E}(X)$ which produces the latent feature $z$ and the generator $\mathcal{G}(z)$, the objective is to make the expectation $\mathbb{E}_{z \sim Q}P(Y|z)$ of the likelihood $P(Y|z)$ to be close to the true probability $P(Y)$, where the probability $Q(z)$ is determined by $\mathcal{E}(X)$ while $P(Y|z)$ is determined by $\mathcal{G}(z)$. 
We then use the Kullback-Leibler (KL) divergence $\mathbb{D}$ from posterior $P(z|Y)$ to $Q(z|X,\mathcal{A})$, written as 
\begin{align}
 \label{equ:target}
 \mathbb{D}_\text{KL}[Q(z|X,\mathcal{A})||P(z|Y)]
 \!=\!\! \mathbb{E}_{z\sim Q}\left[\log \left( \frac{Q(z|X,\mathcal{A})}{P(z|Y)} \right) \right]
\end{align}
to measure the difference between those two distributions. Thus, by minimizing the KL divergence, we evaluate the capacity of the encoder to generate latent variables that are likely to produce the expected target.

Since $P(z|Y)$ is intractable, similar to VAE~\cite{kingma2013auto,blei2017variational}, we rewrite  the KL-divergence from \eqref{equ:target} as
\begin{align}
 &\mathbb{D}_\text{KL}[Q(z|X,\mathcal{A})||P(z|Y)] \nonumber \\ 
 &= \mathbb{E}_{z\sim Q}\left[\log \left( \frac{Q(z|X,\mathcal{A})}{P(Y|z) \cdot P(z)} \right) \right] + \log P(Y) \nonumber \\
 &= \log P(Y) - \left(-\mathbb{E}_{z\sim Q}\left[\log \left( \frac{Q(z|X,\mathcal{A})}{P(Y|z) \cdot P(z)} \right) \right]\right) 
\end{align}
%
where the second term is called evidence lower bound (ELBO) of $\log P(Y)$. 
Considering that the first term is independent of $Q(z|X,\mathcal{A})$, the optimization then focuses on 
\begin{align}
 \text{ELBO}= &-\mathbb{E}_{z\sim Q}\left[\log \left( \frac{Q(z|X,\mathcal{A})}{P(Y|z) \cdot P(z)} \right) \right] \nonumber \\ 
 = &\mathbb{E}_{z\sim Q}\left[\log P(Y|z) \right]-\mathbb{E}_{z\sim Q}\left[\log \left( \frac{Q(z|X,\mathcal{A})}{P(z)} \right) \right]
  \nonumber \\ 
 = &\mathbb{E}_{z\sim Q}\left[\log P(Y|z) \right]-\mathbb{D}_\text{KL}[Q(z|X,\mathcal{A})||P(z)] ~.
\end{align}
%
%
%
Finally, we define the loss function of the generative model as $\mathcal{L}_\text{enc-gen} = -\text{ELBO}$.
The final form of the loss function is written as
\begin{align}
 \label{equ:loss_gmcml_vae}
 \mathcal{L}_\text{enc-gen} = 
 \underbrace{\mathbb{D}[Q(z|X,\mathcal{A}))||P(z)]}_\text{$\mathcal{L}_\text{enc}$} \underbrace{- \mathbb{E}_{z\sim Q}[\log P(Y|z)]}_\text{$\mathcal{L}_\text{gen}$}
\end{align}
where the first term ($\mathcal{L}_\text{enc}$) enforces the encoder to produce latent features which satisfy a Gaussian distribution while the second term ($\mathcal{L}_\text{gen}$) enforces the predicted output from the latent feature fits the expected ground truth.

 \begin{figure}[t]
   \centering
   \includegraphics[width=\linewidth]{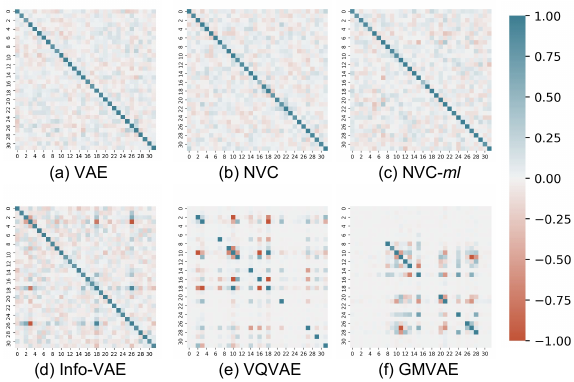}
   \caption{Comparison of covariance matrices computed from the latent features that relies on variational constraint.}
 \label{fig:difference_vae}
 \end{figure}

Although the encoder is optimized by $\mathcal{L}_\text{nebula}$ which enforces clusters and $\mathcal{L}_\text{enc}$ which enforces a Gaussian distribution, it is noteworthy to mention that there is no conflict in optimizing both losses.
In training, our latent feature converges to a single Gaussian distribution while the Nebula loss focuses on forming clusters within this distribution. 
We illustrate this occurrence in \figref{fig:difference_vae} by plotting the covariance matrix of the latent features. This figure show that our convariance matrix is almost identity which behaves similar to VAE~\cite{kingma2013auto}. 
Conversely, other variational clustering approaches like Info-VAE~\cite{zhao2019infovae}, VQ-VAE~\cite{van2017neural} and GMVAE~\cite{dilokthanakul2016deep}, which rely on additional loss functions, discretization or multiple Gaussian distributions, deviate from an identity matrix.

\subsection{Self-supervised metric learning from the anchors}
\label{sec:supervised}

One interesting characteristic of nebula anchors is the capacity to form clusters even under unsupervised or self-supervised settings.
To separate the clusters in the latent space further, we employ the losses involving metric learning. 

Based on these clusters, we label all samples based on the closest nebula anchor. 
%
The labels allow us to apply the self-supervised metric learning on the samples with the Siamese term~\cite{lu2014neighborhood} 
\begin{align}
	\mathcal{L}_\text{pair}(X_i, X_p) = |\mathcal{E}(X_i)-\mathcal{E}(X_p)|^2
	\label{siamese}
\end{align}
so that distances between the samples from the same category become smaller while the distances between the samples from different categories become larger; and, the loss function for triplet learning~\cite{wohlhart15}
\begin{align}
	&\mathcal{L}_\text{triplet}(X_i, X_p, X_n) \nonumber\\ 
	&= \ln\left(\max\left(1,2-\frac{|\mathcal{E}(X_i)-\mathcal{E}(X_n)|^2}{|\mathcal{E}(X_i)-\mathcal{E}(X_p)|^2 + 0.01}\right)\right)
	\label{triplet}
\end{align}
so that samples from the same cluster, \ie the positive pairs $X_i$ and $X_p$, are closer than samples from distinct clusters, \ie negative pairs $X_i$ and $X_n$. We apply these loss functions on all the possible permutations in the batch. 

Consequently, although summing the two losses to $\mathcal{L}_\text{metric}$ does not significantly improve our results, the improvements are consistent throughout all our experiments in the evaluation section. This makes us conclude that they are optional but, at the same time, also valuable to the overall performance.

\subsection{Model optimization}
\label{sec:model}

Based on the previous sections, the final loss function is thus defined as
\begin{align}
 \label{equ:loss_total}
 \mathcal{L}_\text{total} = \mathcal{L}_\text{enc-gen} + \mathcal{L}_\text{nebula} + \mathcal{L}_\text{metric}
\end{align}
%
where the self-supervised metric learning loss term ($\mathcal{L}_\text{metric}$) is optional.
Similar to the optimization in VAE~\cite{kingma2013auto}, the probability density function of the encoder $Q(z|X, a)$ is defined as the Gaussian distribution $N(z|\mu$,$\Sigma)$, where $\mu(X;\theta)$ and $\Sigma(X;\theta)$ are arbitrary deterministic functions.
%
In addition, we use the re-parameterization approach from VAE, explained in~\cite{kingma2013auto}, to optimize the basic generative loss $\mathcal{L}_\text{enc-gen}$.


\section{Experiments}
\label{sec:experiments}

To assess the proposed NVC model, we conduct an extensive evaluation of the proposed method on various applications. From 1D to 3D information, we dive into the following datasets:
%
\begin{enumerate}
    \item \textbf{WMT16}~\cite{bojar2016findings} 
    for 1D language translation of the text sequences;
    \item \textbf{MNIST}~\cite{lecun1998mnist} 
    for 2D image reconstruction;
    \item \textbf{ShapeNet}~\cite{shapenet2015} 
    for 3D semantic completion; 
    \item \textbf{PointNet}~\cite{qi2017pointnet} 
    for 3D point cloud segmentation;
    \item \textbf{Stereo}~\cite{zhang20163d} and \textbf{HOP}~\cite{gao2019variational} 
    for 3D hand pose estimation;
    \item \textbf{NYUv2}~\cite{couprie2013indoor} and \textbf{ScanNet}~\cite{dai2017scannet} 
    for 3D planar reconstruction;
    and,
    \item \textbf{ScanNet}~\cite{dai2017scannet} 
    for 3D semantic scene completion.
\end{enumerate}
The objective is to evaluate the advantage of imposing the nebula anchors on the latent space. Thus, in the following experiments, we assess the difference of with and without NVC where the latent space and nebula anchors are trained in an unsupervised fashion. Moreover, we distinguish NVC from NVC-ML which includes the optional metric learning from \secref{sec:supervised}.

\subsection{1D language translation (WMT16)}

\begin{table}[t]
 \centering
\resizebox{0.99\linewidth}{!}
{\begin{tabular}{l|ccc|c}
	 \toprule
  Method & WMT13 & WMT14 & WMT15 & Variance \\
	 \midrule
  NMT~\cite{wu2016google} (greedy) & 27.1 & - & 27.6 & \textbf{0.19} \\
  NMT~\cite{wu2016google} (beam=10) & 28.0 & - & 28.9 & 0.23 \\
  NMT-GNMT~\cite{wu2016google} & 29.0 & - & 29.9 & 0.30 \\
  -- \textit{with} GMVAE~\cite{aguilera2021regularizing} & 29.8 & - & 30.4 & - \\
  -- \textit{with} \textbf{NVC} & 31.2 & - & 32.4 & 0.25 \\
  -- \textit{with} \textbf{NVC-ML} & \textbf{33.4} & - & \textbf{34.9} & 0.31 \\
	 \midrule
  FusedBert~\cite{zhu2019incorporating} & - & 30.8 & - & - \\
  -- \textit{with} GMVAE~\cite{aguilera2021regularizing} & - & 31.2 & - & - \\
  -- \textit{with} \textbf{NVC} & - & 31.6 & - & 0.19 \\
  -- \textit{with} \textbf{NVC-ML} & - & 32.1 & - & 0.26 \\
    \midrule
  Transformer-Rep~\cite{takase2021rethinking} & - & 33.9 & - & - \\
  -- \textit{with} GMVAE~\cite{aguilera2021regularizing} & - & 33.9 & - & - \\
  -- \textit{with} \textbf{NVC} & - & 34.1 & - & 0.31 \\
  -- \textit{with} \textbf{NVC-ML} & - & \textbf{34.2} & - & 0.33 \\
	 \bottomrule
 \end{tabular}
 }
 \caption{Evaluation on the WMT German-English translation~\cite{bojar2016findings}, performance is reported in BLEU score~\cite{takase2021rethinking}. \label{tab:wmt}}
\end{table}

Neural machine translation system usually rely on sequence-to-sequence models such as~\cite{sutskever2014sequence, britz2017massive, cho2014learning}. These models embed 1D input sentences by means of an encoder; then, a recurrent model -- typically a Long-Short Term Memory (LSTM)~\cite{hochreiter1997long} network -- operates on the latent space; and finally, a decoder processes the embedded representation to obtain an output translation and to capture the long-range dependencies in the sentences. 

We propose an experiment based on the WMT German-English dataset~\cite{bojar2016findings}. One of the baseline architectures is a 4-layer Neural Machine Translation (NMT) model~\cite{wu2016google} with LSTM units. We apply our NVC model on the 1024-dimensional embedding of the last GNMT layer. NVC is applied with and without metric learning, \ie only with nebula anchors, to train the latent variables in a fully unsupervised way. 

Due to the popularity of transformers, we also investigate using FusedBert~\cite{zhu2019incorporating} and TransformerRep~\cite{takase2021rethinking} as our baseline architectures. 
For the transformers, we apply NVC on the fused latent feature of the BERT-encoder attention and the self-attention.

\tabref{tab:wmt} shows the advantages of our NVC approach, demonstrating its capacity to generalize even when applied on recurrent models like
NMT~\cite{wu2016google}, FusedBert~\cite{zhu2019incorporating} and TransformerRep~\cite{takase2021rethinking}.
The results from the other methods are taken from~\cite{luong17}.

Utilizing the same baseline architectures, we also compare the our NVC against GMVAE~\cite{dilokthanakul2016deep} in order to quantify the difference between the two clustering techniques. 
\tabref{tab:wmt} demonstrates that GMVAE~\cite{dilokthanakul2016deep} marginally improves the machine translation models while the proposed method reveals a clear improvement.

\subsection{2D image reconstruction (MNIST)}

We also evaluate the MNIST~\cite{lecun1998mnist} dataset to demonstrate that training with NVC performs as theoretically expected, especially in terms of the learned digit-aware sub-manifolds in the latent space centered on the anchors.
The MNIST dataset includes hand written digits characters from 10 categories which are evaluated based on the reconstruction precision of the auto-encoder.

We adopt an auto-encoder as the basic architecture with three fully-connected layers for each of the encoder and the decoder.
Notably, we train the NVC with self-supervised metric learning without the categorical label of the samples.
To compare against the variational inference models without the latent anchors, one hidden layer is added to produce the latent feature to train a VAE, a DVAE and a CVAE.

In \tabref{tab:mnist_rec}, we compare a series of variational inference models such as VAE~\cite{kingma2013auto}, DVAE~\cite{im2017denoising} and CVAE~\cite{Sohn2015cgm} with and without NVC. 
Using the reconstruction metrics proposed in~\cite{eigen2015predicting}, this table demonstrates the effectiveness of the proposed variational coder in improving the models.
We also include the experiments where the anchors are trained by assigning the ground truth categorical labels during the metric learning which is referred as \textit{Supervised} in \tabref{tab:mnist_rec}. It shows that our proposed self-supervised learning, resulting in having digit-related embeddings, is helpful to improve the baseline methods.

\begin{table}[t]
 \centering
 {\begin{tabular}{l|l|cccc}
	\toprule	
  & Method & $rel$ & $\delta_1$ & $\delta_2$ & $\delta_3$ \\
	\midrule	
  \multirow{7}{*}{\rotatebox[origin=c]{90}{\centering \scriptsize{Unsupervised}}} & Auto-encoder & 0.191 & 78.6\% & 85.3\% & 91.9\% \\
	       & DVAE~\cite{im2017denoising} & 0.188 & 80.2\% & 86.8\% & 92.8\% \\
	       & -- \textit{with} \textbf{NVC} & 0.162 & 82.8\% & 89.2\% & 95.0\% \\
	       & -- \textit{with} \textbf{NVC-ML} & 0.147 & 84.5\% & 92.7\% & 96.2\% \\
	       & VAE~\cite{kingma2013auto} & 0.169 & 82.4\% & 91.3\% & 95.2\% \\
	       & -- \textit{with} \textbf{NVC} & 0.138 & 84.3\% & 92.6\% & 96.1\% \\
	       & -- \textit{with} \textbf{NVC-ML} & \textbf{0.133} & \textbf{85.3\%} & \textbf{93.8\%} & \textbf{97.2\%} \\
	\midrule	
  \multirow{4}{*}{\rotatebox[origin=c]{90}{\centering \scriptsize{Supervised}}} & DVAE \textit{with} \textbf{NVC-ML} & 0.142 & 85.3\% & 92.8\% & 96.9\% \\
	     & VAE \textit{with} \textbf{NVC-ML} & 0.141 & 85.1\% & 92.5\% & 96.3\% \\
	     & CVAE~\cite{Sohn2015cgm} & 0.165 & 82.9\% & 91.7\% & 95.4\% \\
	     & -- \textit{with} \textbf{NVC-ML} & \textbf{0.116} & \textbf{87.6\%} & \textbf{94.3\%} & \textbf{98.6\%} \\
	\bottomrule	
 \end{tabular}
 }
 \caption{Evaluation of the reconstruction on MNIST~\cite{lecun1998mnist} for different variational models. We use 10 anchors in the latent space. The term $rel$ is the absolute relative error while the threshold accuracy $\delta_i$ is described in Adabins~\cite{bhat2021adabins}. \label{tab:mnist_rec}
 }
\end{table}


%

\begin{figure}[!ht]
\captionsetup[subfigure]{justification=centering}
 \centering
   \begin{subfigure}[b]{0.135\linewidth}
    \centering
    \centerline{\includegraphics[width=1\linewidth]{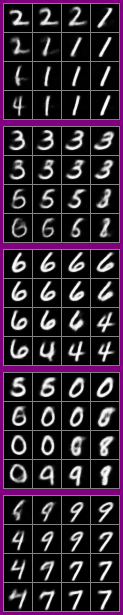}}
    \caption{5}
   \end{subfigure}
   \begin{subfigure}[b]{0.27\linewidth}
    \centering
    \centerline{\includegraphics[width=1\linewidth]{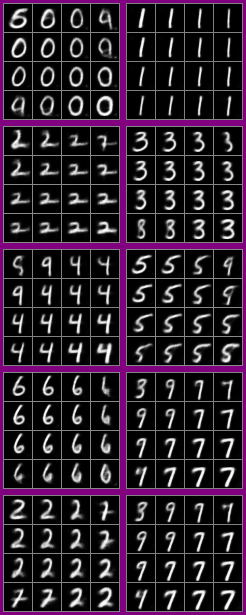}}
    \caption{10}
   \end{subfigure}
   \begin{subfigure}[b]{0.54\linewidth}
    \centering
    \centerline{\includegraphics[width=1\linewidth]{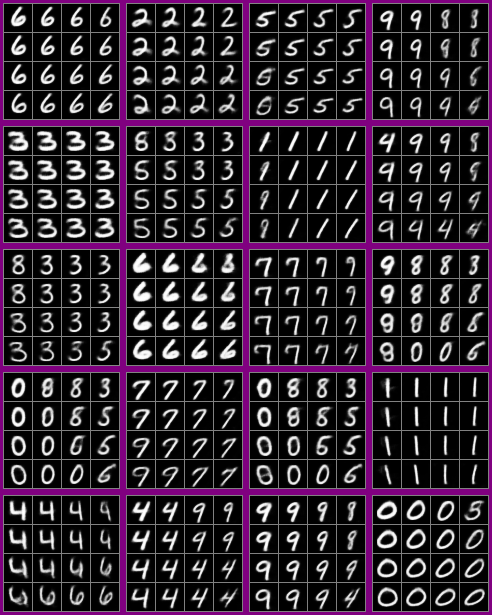}}
    \caption{20}
   \end{subfigure}
\caption{Visualization of the manifold centered on the nebula anchors learned with different numbers of anchors.}
\label{fig:exp_mnist_orders}
\end{figure}

For this experiment, it is noteworthy to mention that the simplicity of the auto-encoder helps us generate insights on the characteristics inferred by the learned nebula anchors.
\figref{fig:exp_mnist_orders} shows that different numbers of nebula anchors perform well in learning meaningful manifolds according to the hidden information such as the digit labels. 
When using five nebula anchors, the five manifolds already include all 10 categories ranging from 0 to 9.
Increasing to 10 anchors reveal that every anchor is surrounded by each of the 10 digits. If we increase further to 20 anchors, they are not only separate by the digits but also by the font styles.
Later in \secref{sec:num_anchors}, we numerically evaluate the number of anchors from different datasets.

\begin{figure}[!hb]
\centering
\includegraphics[width=0.9\linewidth]{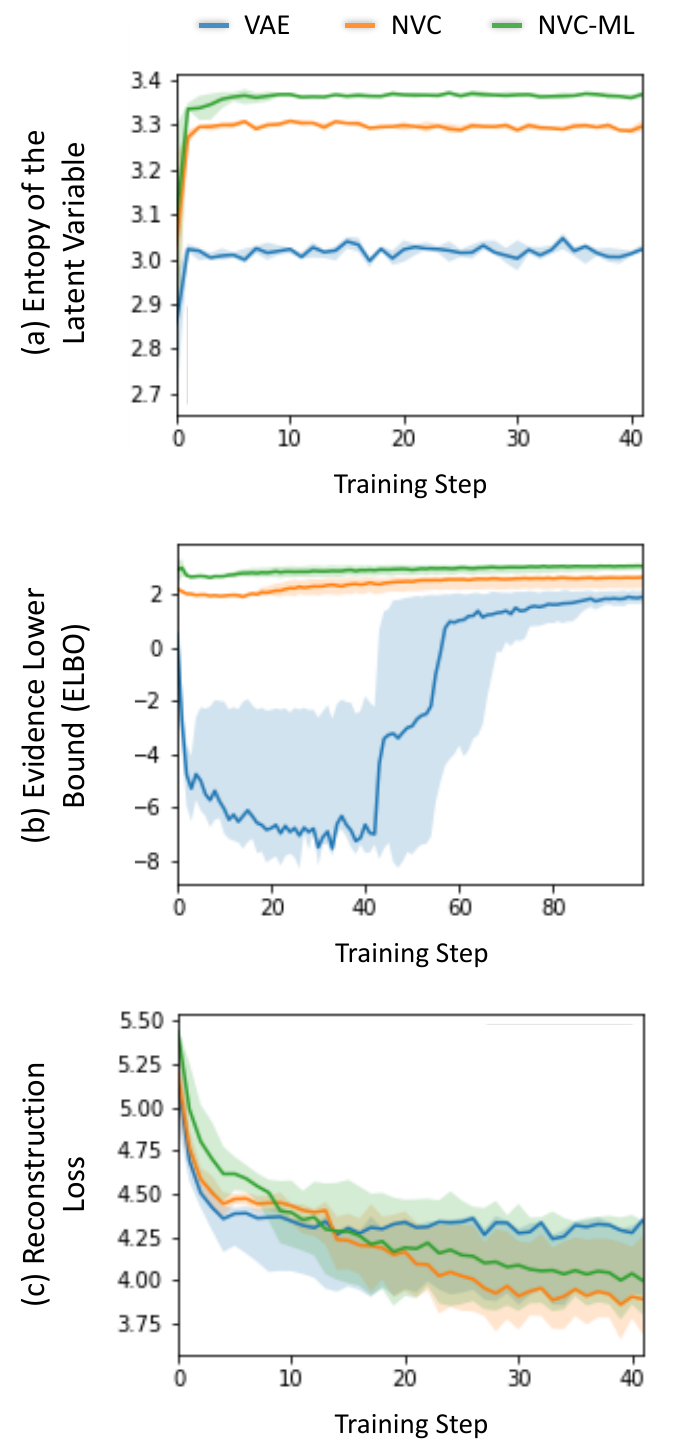}
\caption{
Influence of the nebula anchors on 
(a) the entropy of the latent variables, 
(b) the evidence lower bound (ELBO) 
and 
(c) the reconstruction loss during training.
}
\label{fig:exp_mnist_training_linechart}
\end{figure}


In addition, we investigate the behavior of the latent space during training. By comparing VAE with and without the proposed methods, \ie NVC and NVC-M, \figref{fig:exp_mnist_training_linechart}~(a) plots the entropy of the latent variables at each training step. Based on this figure, the entropy with NVC is larger while adding the self-supervised metric learning improves it further. 
Note that the higher entropy implies that clusters formed are more separated.
Hence, the encoder optimized with the nebula anchors produces a more informative embedding.
Although this also implies that the higher entropy deviates from the Gaussian distribution, we visualize in \figref{fig:difference_vae} that the optimized distribution remains similar to Gaussian.

To validate the optimization target $\mathcal{L}_\text{enc}$ in (\ref{equ:loss_gmcml_vae}), we also visualize the evidence lower bound (ELBO) at each training step in \figref{fig:exp_mnist_training_linechart}~(b). 
Here, it becomes evident that the ELBO with the proposed method is raised compared to standard VAE so that its gap from the true posterior (which is always larger than ELBO) becomes smaller.
Consequently, the reconstruction loss converges to a smaller error with optimized with the proposed Nebula anchors as illustrated in \figref{fig:exp_mnist_training_linechart}~(c). 



\begin{table}[t]
 \centering
\resizebox{1.\linewidth}{!}
 {\begin{tabular}{l|ccccc}
	\toprule
	Methods & 10 & 16 & 20 & 30 & 120 \\
	\midrule
    GMVAE~\cite{dilokthanakul2016deep} & 88.54 & 91.26 & 96.92 & 93.22 & 89.70 \\
    VQ-VAE~\cite{van2017neural} & 72.32 & 81.45 & 82.60 & 94.22 & 96.89 \\
	\midrule
	NVC & \textbf{98.21} & \textbf{97.79} & \textbf{97.51} & \textbf{97.19} & \textbf{97.03} \\
	\bottomrule	
 \end{tabular}
 }
 \caption{Compares different number of anchors (or clusters) on the classification accuracy evaluated on MNIST~\cite{lecun1998mnist}.} \label{tab:mnist_cls}
\end{table}

Moreover, we also compare different clustering approaches, varying the number of anchors (or clusters). 
\tabref{tab:mnist_cls} illustrates the adaptability of the proposed method to acquire consistent results across different anchor sizes. This is in contrast to the other methods~\cite{dilokthanakul2016deep, van2017neural} where they peak at only specific number of classes.

\subsection{3D volumetric completion (ShapeNet)}
\label{sec:voxel_completion}

Adapting the evaluation strategy from 3D-RecGAN~\cite{yang2018dense}, we use ShapeNet~\cite{shapenet2015} to generate the training and test data for 3D object completion, wherein each reconstructed object surface is paired with a corresponding ground truth voxelized shape with a size of $64\times64\times64$. 
%
%
The dataset comprises of four object classes: \textit{bench}, \textit{chair}, \textit{couch} and \textit{table}. Note that \cite{yang2018dense} prepared an evaluation for both synthetic and real input data.

\begin{table}[b]
\centering
{\begin{tabular}{l|l|cccc|c}
	\toprule	
	 & Method & bench & chair & couch & table & \emph{Avg.} \\
	\midrule
	\multirow{12}{*}{\rotatebox[origin=c]{90}{\centering {Synthetic~\cite{shapenet2015}}}}  & Varley et.al~\cite{varley2017shape} & 65.3 & 61.9 & 81.8 & 67.8 & 69.2 \\
		       & 3D-EPN~\cite{dai2017shape} & 75.8 & 73.9 & 83.4 & 77.2 & 77.6 \\
		       & Han et.al~\cite{han2017high} & 54.4 & 46.9 & 48.3 & 56.0 & 51.4 \\
		       & 3D-AE~\cite{yang2018dense} & 73.3 & 73.6 & 83.2 & 75.0 & 76.3 \\
		       & -- \textit{with}  \textbf{NVC} & 74.9 & 73.9 & 84.8 & 76.4 & 77.5 \\
		       & -- \textit{with}  \textbf{NVC-ML} & 76.5 & 74.4 & 85.6 & 77.6 & 78.6 \\
		       & 3D-RecGAN~\cite{yang2018dense} & 74.5 & 74.1 & 84.4 & 77.0 & 77.5 \\
		       & -- \textit{with}  \textbf{NVC} & 76.7 & 78.4 & 90.0 & 81.4 & 81.6 \\
		       & -- \textit{with}  \textbf{NVC-ML} & 78.2 & 79.1 & 92.7 & 82.8 & 83.2 \\
		       & ForkNet~\cite{wang2019forknet} & 79.1 & 80.6 & 92.4 & 84.0 & 84.1 \\
		       & -- \textit{with}  \textbf{NVC} & 80.2 & 82.1 & 92.9 & 85.3 & 85.1 \\
		       & -- \textit{with}  \textbf{NVC-ML} & \textbf{81.9} & \textbf{83.5} & \textbf{93.4} & \textbf{86.0} & \textbf{86.2} \\
	\midrule
		    \multirow{11}{*}{\rotatebox[origin=c]{90}{\centering Real~\cite{yang2018dense}}}
		       & Han et.al~\cite{han2017high} & 18.4 & 14.8 & 10.1 & 12.6 & 14.0 \\ 
		       & 3D-AE~\cite{yang2018dense} & 23.1 & 17.8 & 10.7 & 14.8 & 16.6 \\ 
		       & -- \textit{with} \textbf{NVC} & 23.6 & 18.1 & 10.7 & 16.1 & 17.1 \\ 
		       & -- \textit{with}  \textbf{NVC-ML} & 25.0 & 19.2 & 12.5 & 16.9 & 18.4 \\ 
		       & 3D-RecGAN~\cite{yang2018dense} & 23.0 & 17.4 & 10.9 & 14.6 & 16.5 \\ 
		       & -- \textit{with} \textbf{NVC} & 27.9 & 19.1 & 13.6 & 17.8 & 19.6 \\ 
		       & -- \textit{with} \textbf{NVC-ML} & 29.2 & 19.8 & 14.6 & 18.4 & 20.5 \\
		       & ForkNet~\cite{wang2019forknet} & 32.7 & 24.1 & 15.9 & 22.5 & 23.8 \\ 
		       & -- \textit{with} \textbf{NVC} & 34.1 & 25.0 & 16.4 & 24.2 & 24.9 \\
		       & -- \textit{with} \textbf{NVC-ML} & \textbf{34.9} & \textbf{26.8} & \textbf{17.8} & \textbf{25.5} & \textbf{26.3} \\
	\bottomrule
	\end{tabular}
}
\caption{Evaluation of the object completion in terms of IoU (in $\%$) on ShapeNet~\cite{shapenet2015}. The resolution of Varley et.al~\cite{varley2017shape} and 3D-EPN~\cite{dai2017shape} is $32 \times 32 \times 32$ while $64 \times 64 \times 64$ for the others. \label{tab:shapenet_iou}}
\end{table}

\begin{table*}[t]
\centering
\resizebox{\textwidth}{!}
{\begin{tabular}{l|cccccccccccccccc|c}
	\toprule	
	Method & aero & bag & cap & car & chair & earpod & guitar & knife & lamp & laptop & motor & mug & pistol & rocket & skateboard & table & \emph{Avg.} \\
	\midrule 
	Wu~\etal~\cite{wu2014interactive} & 63.2 & - & - & - & 73.5 & - & - & - & 74.4 & - & - & - & - & - & - & 74.8 & - \\
	Yi~\etal~\cite{yi2016scalable} & 81.0 & 78.4 & 77.7 & 75.7 & 87.6 & 61.9 & \textbf{92.0} & 85.4 & \textbf{82.5} & \textbf{95.7} & \textbf{70.6} & 91.9 & \textbf{85.9} & 53.1 & 69.8 & 75.3 & 81.4 \\
	3DCNN~\cite{qi2016volumetric} & 75.1 & 72.8 & 73.3 & 70.0 & 87.2 & 63.5 & 88.4 & 79.6 & 74.4 & 93.9 & 58.7 & 91.8 & 76.4 & 51.2 & 65.3 & 77.1 & 79.4 \\
	SGPN~\cite{wang2018sgpn} & 80.4 & 78.6 & 78.8 & 71.5 & 88.6 & 78.0 & 90.9 & 83.0 & 78.8 & 95.8 & 77.8 & 93.8 & 87.4 & 60.1 & 92.3 & 89.4 & 85.8 \\
	SSCNN~\cite{yi2017syncspeccnn} & 81.6 & 81.7 & 81.9 & 75.2 & 90.2 & 74.9 & 93.0 & 86.1 & 84.7 & 95.6 & 66.7 & 92.7 & 81.6 & 60.6 & 82.9 & 82.1 & 84.7 \\
	-- \textit{with} \textbf{NVC-ML} & 77.8 & 75.6 & 77.0 & 72.1 & 89.5 & 64.2 & 91.9 & 82.3 & 76.2 & 94.1 & 60.5 & 94.2 & 78.0 & 54.4 & 67.9 & 80.0 & 77.2 \\
	PointNet++~\cite{qi2017pointnet++} & 82.4 & 79.0 & 87.7 & 77.3 & 90.8 & 71.8 & 91.0 & 85.9 & 83.7 & 95.3 & 71.6 & 94.1 & 81.3 & 58.7 & 76.4 & 82.6 & 85.1 \\
	PointNet~\cite{qi2017pointnet} & 83.4 & 78.7 & 82.5 & 74.9 & 89.6 & \textbf{73.0} & 91.5 & 85.9 & 80.8 & 95.3 & 65.2 & \textbf{93.0} & 81.2 & 57.9 & 72.8 & 80.6 & 83.7 \\
	-- \textit{with} \textbf{NVC-ML} & 84.1 & 81.6 & 82.7 & 76.5 & 89.8 & 71.6 & 91.8 & 86.1 & 81.7 & 95.7 & 66.3 & 92.2 & 82.3 & 61.0 & 73.2 & 81.7 & 84.4 \\
	3D-GCN~\cite{lin2020convolution} & 83.1 & \textbf{84.0} & 86.6 & 77.5 & 90.3 & \textbf{74.1} & 90.9 & 86.4 & 83.8 & 95.6 & 66.8 & 94.8 & 81.3 & 59.6 & 75.7 & 82.8 & 85.1 \\
	-- \textit{with} \textbf{NVC-ML} & \textbf{83.5} & 83.1 & \textbf{91.4} & \textbf{79.6} & \textbf{92.9} & 70.3 & \textbf{97.4} & \textbf{87.1} & \textbf{87.4} & \textbf{97.0} & \textbf{78.0} & \textbf{97.2} & \textbf{84.4} & \textbf{61.6} & \textbf{82.3} & \textbf{87.9} & \textbf{86.3} \\
	\bottomrule
\end{tabular}
}
\caption{Evaluation of the semantic point cloud segmentation on ShapeNet~\cite{shapenet2015}. The results are reported in terms of mIoU as global average as well as for each of the 16 classes of the dataset.\label{tab:pointnet}}
\end{table*}

\subsubsection{Synthetic data}

We perform two evaluations in \tabref{tab:shapenet_iou}. The first is a single category test~\cite{yang2018dense} such that each category is trained and tested separately while the second considers the categories in order to label the voxels. 
We compare our results against \cite{dai2017shape,han2017high,varley2017shape,yang2018dense,wang2019forknet} by applying the nebula anchors while learning the latent feature. The results are reported in IoU with a 3D resolution of $64 \times 64 \times 64$. By introducing the anchors, the IoU of 3D-AE~\cite{yang2018dense}, 3D-RecGAN~\cite{yang2018dense} and ForkNet~\cite{wang2019forknet} are improved from 76.3\%, 77.5\% and 84.1\% to 77.5\%, 81.6\% and 85.1\%, respectively. With metric learning, the result obtained from ForkNet~\cite{wang2019forknet} achieves the best performance of 86.2\% among all the approaches.

\subsubsection{Real data}

Since both 3D-RecGAN~\cite{yang2018dense} and ForkNet~\cite{wang2019forknet} evaluated the real scans~\cite{yang2018dense} captured by Kinect, 
we also evaluate them with and without the proposed method. 

The single category test in \tabref{tab:shapenet_iou} suggested by 3D-RecGAN~\cite{yang2018dense} shows that, by simply optimizing the latent features using our approach (\ie nebula anchor with metric learning), the completion IoU is improved by 4\% on average while 2.5\% on ForkNet~\cite{wang2019forknet}.
In addition, the results in IoU also show that, even without the metric learning, the nebula anchors help improve 3D-RecGAN by 3.1\% and ForkNet by 1.1\%.

\begin{figure}[t]
\captionsetup[subfigure]{justification=centering}
 \centering
   \begin{subfigure}[b]{1.\linewidth}
    \centering
    \centerline{\includegraphics[width=1\linewidth]{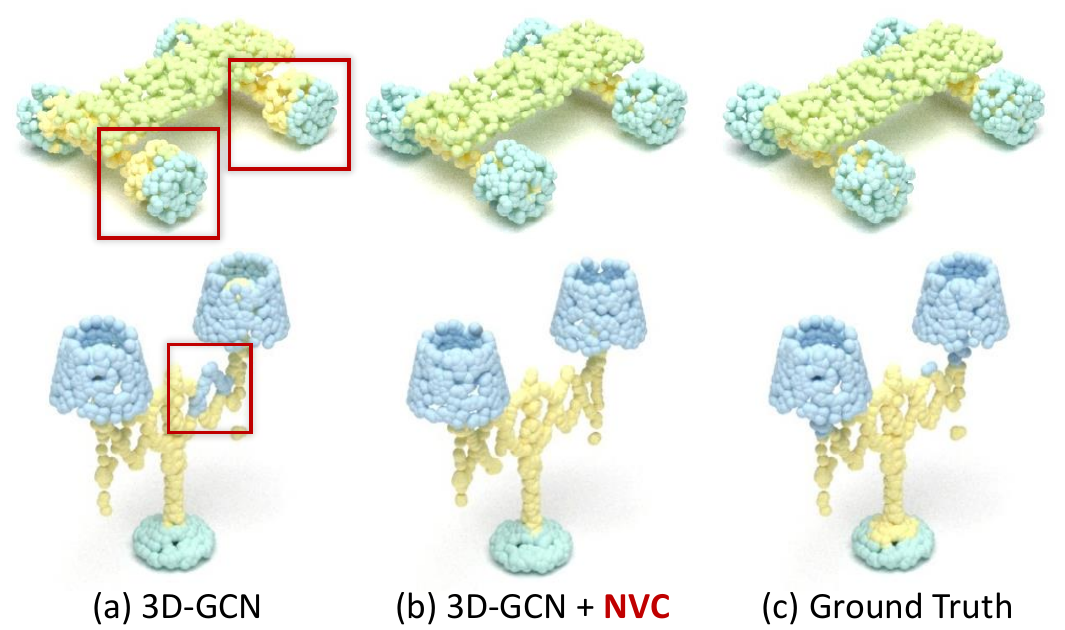}}
   \end{subfigure}
   \\
\caption{Comparison of part segmentation from 3D-GCN~\cite{lin2020convolution} with and without the proposed NVC optimization.}
\label{fig:exp_pointnet_pred}
\end{figure}

\subsection{3D point cloud segmentation (PointNet)}

We evaluated the proposed method in a 3D semantic segmentation task for point clouds. In particular, we focus on the dataset proposed by PointNet~\cite{qi2017pointnet}, which is a subset of ShapeNet including 16 categories and a total of 16,881 point clouds. In this case, we applied our NVC model to the network architecture proposed in PointNet, and again trained the self-supervised metric learning using the class labels of each point cloud. 

\tabref{tab:pointnet} shows the semantic segmentation results of NVC against the state of the art in terms of mean-Intersection-over-Union (mIoU), \ie the standard metric for this dataset~\cite{qi2017pointnet}. The table shows that NVC is beneficial to improve the performance of PointNet, and achieve the best overall result and on most individual categories.

In addition, we show some qualitative results in \figref{fig:exp_pointnet_pred}, comparing the segmentation results from 3D-GCN~\cite{lin2020convolution} with those reported by our method. From the figure, we can see that NVC yields a more accurate segmentation, allowing to better distinguish the borders between different object parts.

\begin{table}[!b]
\centering
\resizebox{0.99\linewidth}{!}
{
	\begin{tabular}{c|l|ccc}

	\toprule
		 & Method & AUC & EPE Median & EPE Mean \\
	\midrule
	\multirow{5}{*}{\rotatebox[origin=c]{90}{\centering Stereo~\cite{zhang20163d}}}
		     & CHPR~\cite{sun2015cascaded} & 0.839 & - & - \\ 
		     & GANeratedHands~\cite{GANeratedHands_CVPR2018} & 0.965 & - & - \\
		     & PosePrior~\cite{zimmermann2017learning} & 0.948 & 9.543 & 11.064 \\ 
		     & VO-Hand~\cite{gao2019variational} (cluster) & 0.984 & 7.606 & 8.943 \\ 
		     & -- \textit{with} \textbf{NVC} & \textbf{0.986} & \textbf{7.511} & \textbf{8.897} \\
	\midrule
	\multirow{5}{*}{\rotatebox[origin=c]{90}{\centering HOP~\cite{gao2019variational}}}
		  & PosePrior~\cite{zimmermann2017learning} & 0.534 & 19.728 & 30.860 \\
		     & VO-Hand~\cite{gao2019variational} (triplet) & 0.597  & 15.901 & 27.326 \\
		     & -- \textit{with} \textbf{NVC} & \textbf{0.623}  & \textbf{13.935} & \textbf{26.405} \\
		     & VO-Hand~\cite{gao2019variational} (cluster) & 0.583 & 16.741 & 28.018 \\
		     & -- \textit{with} \textbf{NVC} & 0.599 & 16.063 & 27.729 \\
	\bottomrule
	\end{tabular}
}
\caption{Evaluation on Stereo~\cite{zhang20163d} and HOP~\cite{gao2019variational} with the baseline approach from VO-Hand~\cite{gao2019variational} trained with unsupervised clustering (cluster)~\cite{gao2019variational} or triplet learning (triplet). The endpoint errors (EPE) are in millimeters. 
\label{tab:vohand}
}
\end{table}

\begin{table*}[t]
\centering
\begin{tabular}{l|l|cccccccccccc}
	\toprule	
	 & Method & 0.05 & 0.10 & 0.15 & 0.20 & 0.25 & 0.30 & 0.35 & 0.40 & 0.45 & 0.50 & 0.55 & 0.60 \\
	\midrule
	\multirow{5}{*}{\rotatebox[origin=c]{90}{\centering Pixel}}
    & MWS~\cite{furukawa2009manhattan} & 2.40 & 8.02 & 13.70 & 18.06 & 22.42 & 26.22 & 28.65 & 31.13 & 32.99 & 35.14 & 36.82 & 38.09 \\
    & NYU-Toolbox~\cite{silberman2012indoor} & 3.97 & 11.56 & 16.66 & 21.33 & 24.54 & 26.82 & 28.53 & 29.45 & 30.36 & 31.46 & 31.96 & 32.34 \\
    & PlaneNet~\cite{liu2018planenet} & 22.79 & 42.19 & 52.71 & 58.92 & 62.29 & 64.31 & 65.20 & 66.10 & 66.71 & 66.96 & 67.11 & 67.14 \\
    & PlanarRecon~\cite{yu2019single} & 30.59 & 51.88 & 62.83 & 68.54 & 72.13 & 74.28 & 75.38 & 76.57 & 77.08 & 77.35 & 77.54 & 77.86 \\
    & -- \textit{with} \textbf{NVC} & \textbf{33.45} & \textbf{54.91} & \textbf{65.74} & \textbf{71.05} & \textbf{75.43} & \textbf{77.61} & \textbf{78.10} & \textbf{79.91} & \textbf{80.12} & \textbf{80.67} & \textbf{81.66} & \textbf{81.92} \\
	\midrule
	\multirow{5}{*}{\rotatebox[origin=c]{90}{\centering Plane}}  
    & MWS~\cite{furukawa2009manhattan} & 1.69 & 5.32 & 8.84 & 11.67 & 14.40 & 16.97 & 18.71 & 20.47 & 21.68 & 23.06 & 24.09 & 25.13 \\
    & NYU-Toolbox~\cite{silberman2012indoor} & 3.14 & 9.21 & 13.26 & 16.93 & 19.63 & 21.41 & 22.69 & 23.48 & 24.18 & 25.04 & 25.50 & 25.85 \\
    & PlaneNet~\cite{liu2018planenet} & 15.78 & 29.15 & 37.48 & 42.34 & 45.09 & 46.91 & 47.77 & 48.54 & 49.02 & 49.33 & 49.53 & 49.59 \\
    & PlanarRecon~\cite{yu2019single} & 22.93 & 40.17 & 49.40 & 54.58 & 57.75 & 59.72 & 60.92 & 61.84 & 62.23 & 62.56 & 62.76 & 62.93 \\
    & -- \textit{with} \textbf{NVC} & \textbf{26.98} & \textbf{44.22} & \textbf{53.12} & \textbf{58.65} & \textbf{62.30} & \textbf{64.63} & \textbf{65.07} & \textbf{66.18} & \textbf{67.11} & \textbf{67.68} & \textbf{67.91} & \textbf{68.11} \\
	\bottomrule
	\end{tabular}
\caption{Evaluation of the depth estimation on ScanNet~\cite{dai2017scannet} incorporating 3D plane estimations for the indoor scenes. The pixel and plane recalls are reported according to threshold of depth difference.
\label{tab:depth_estimation}}
\end{table*}

\subsection{3D hand pose estimation (Stereo/HOP)}

To compare our work with VO-Hand~\cite{8770083} which also uses trainable anchors to define the cluster centers, we tried to apply 5 anchors on the HOP~\cite{8770083} dataset, 7 on RHD~\cite{zimmermann2017learning}, and 10 on GANeratedHands~\cite{GANeratedHands_CVPR2018} and Stereo~\cite{zhang20163d}.
In \cite{8770083}, a high number of clusters tends to reduce its advantages while a small number of clusters brings the network back to the standard variational inference model. 

\tabref{tab:vohand} shows the result of the unsupervised training of the latent variables on those datasets, where the proposed nebula anchors improves the performance of VO-hand~\cite{gao2019variational} on the HOP~\cite{gao2019variational} dataset where the hands are occluded by grasped objects from 0.597 to 0.623 in terms of AUC. In the Stereo~\cite{zhang20163d} dataset where hands are not occluded, we also improve the performance in AUC from 0.984 to 0.986.
Note that the pose estimation in Stereo~\cite{zhang20163d} is easier than HOP~\cite{gao2019variational} because the former does not include any hand occlusion; thus, the results are saturated. This explains why the improvement between VO-hand~\cite{gao2019variational} and our work is marginal in Stereo~\cite{zhang20163d}.

\begin{figure}[!t]
 \centering
 \includegraphics[width=\linewidth]{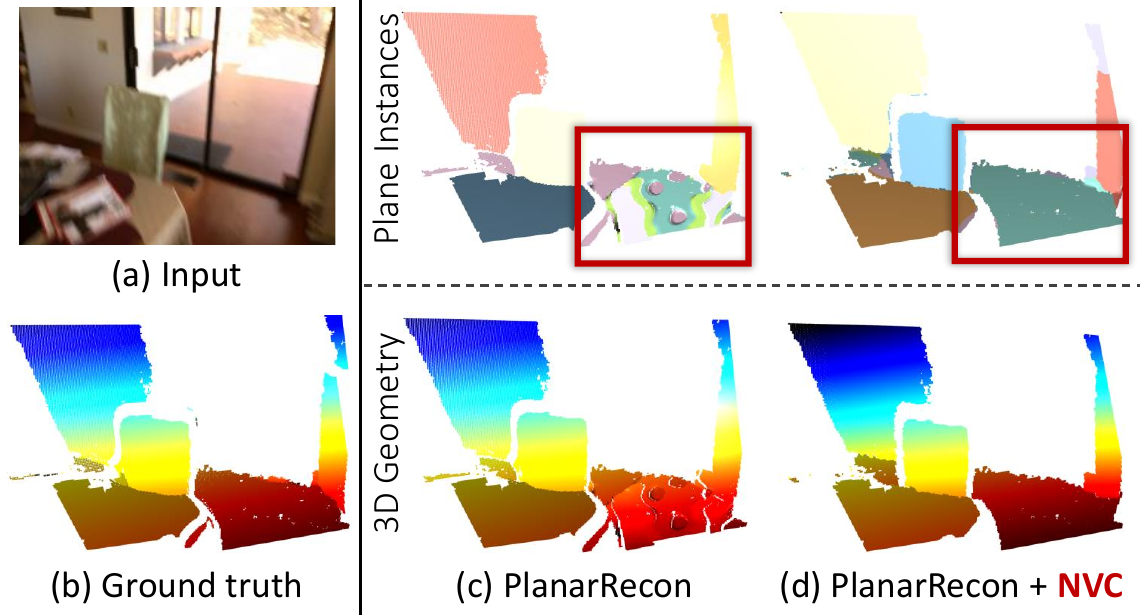}
 \caption{
  Comparison of planar reconstruction from PlanarRecon~\cite{YuZLZG19} with and without the proposed NVC optimization.}
 \label{fig:planar}
\end{figure}

\subsection{3D planar reconstruction (ScanNet)}

The ScanNet dataset~\cite{dai2017scannet} provides real RGB images and their corresponding depth captured by depth cameras.
Aiming at simplifying the 3D reconstruction using large planes, PlanarRecon~\cite{YuZLZG19} further fits 3D planes to the consolidated 3D point cloud which is back-projected from the depth images using the camera's intrinsic parameters. While constructing the 3D geometries in instance-level planes, PlanarRecon~\cite{YuZLZG19} also incorporates the semantic annotations from ScanNet. The resulting dataset
contains 50,000 training and 760 testing images with a resolution of 256$\times$192. 

We evaluate our approach by optimizing the latent feature of PlanarRecon~\cite{yu2019single} using nebula anchors on both ScanNet~\cite{dai2017scannet} and NYUv2~\cite{couprie2013indoor} dataset. Pixel and plane recalls are reported in \tabref{tab:depth_estimation}, where the introduced nebula anchor improves the performance on all threshold of depth difference. By calculating with pixel-wise absolute difference (rel) in \tabref{tab:ablation_anchors} on NYUv2~\cite{couprie2013indoor} dataset, the performance of PlanarRecon~\cite{yu2019single} is improved from 0.134 to 0.126 with the help of adding anchors in latent space.
This improvement is illustrated in \figref{fig:planar} where the blurry RGB input makes PlanarRecon~\cite{yu2019single} hard to reconstruct the floor as a single plane, while the added nebula anchor helps to reconstruct the floor as a whole piece.
Additionally, the best performance is also achieved as 0.125 using our proposed metric learning.

\begin{figure*}[t]
\centering
\includegraphics[width=0.99\linewidth]{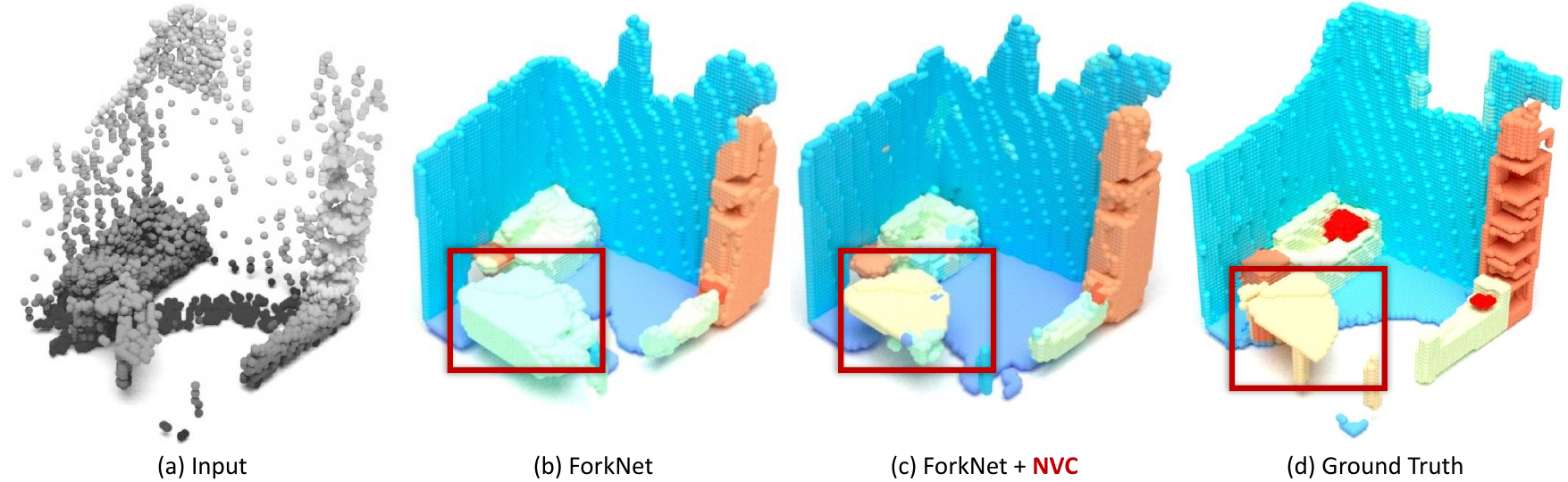}
\caption{
Comparison of semantic scene completion from ForkNet~\cite{wang2019forknet} with and without the proposed NVC optimization.}
\label{fig:semantic_completion}
\vspace{15pt}
\end{figure*}

\begin{table*}[!t]
\centering
\resizebox{0.9\textwidth}{!}
{
\begin{tabular}{l|ccccccccccc|c}
\toprule	
 \multicolumn{1}{c}{Method} 
 & ceil. & floor & wall & win. & chair & bed & sofa & table & tvs & furn. & objs & \emph{Avg.} \\
\midrule 
 ForkNet~\cite{wang2019forknet} & 5.2 & 22.5 & 12.3 & 0.0 & 9.2 & 5.7 & 18.2 & 14.9 & 0.1 & 12.6 & 1.8 & 9.3 \\
 ForkNet~\cite{wang2019forknet}+NVC & 10.2 & 29.3 & 18.2 & 3.3 & 14.5 & 11.9 & 25.1 & 17.9 & 4.7 & 16.0 & 3.9 & 14.1 \\
 ScanComplete~\cite{dai2018scancomplete} & 16.4 & 39.3 & 35.0 & 1.8 & 20.4 & 3.8 & 11.2 & 27.7 & 0.6 & 13.2 & 7.8 & 16.1 \\
 SCFusion~\cite{wu2020scfusion} & 12.8 & 32.9 & 26.5 & 9.6 & 22.5 & 20.7 & 26.4 & 21.0 & 7.4 & 19.2 & 8.6 & 18.9 \\
\bottomrule
\end{tabular}
}
\caption{Evaluation of the semantic completion on CompleteScanNet~\cite{wu2020scfusion}. The results are reported in terms of IoU at a resolution of $64 \times 64 \times 64$.
 \label{tab:completescannet}
}
\vspace{20pt}
\end{table*}

\subsection{3D semantic completion (CompleteScanNet)}

We also evaluate on the 3D semantic completion from the  CompleteScanNet~\cite{wu2020scfusion} dataset which is built from the ScanNet dataset~\cite{dai2017scannet}.
Here, we apply the proposed nebula anchor in the latent space of ForkNet~\cite{wang2019forknet} where the variational constraint is already used. The results are compared to original ForkNet~\cite{wang2019forknet}, ScanComplete~\cite{dai2018scancomplete} and SCFusion~\cite{wu2020scfusion} in terms of IoU on 11 categories. 

By incorporating NVC, \tabref{tab:completescannet} demonstrates a significant improvement on ForkNet increasing the IoU from 9.3\% to 14.1\%. 
We illustrate an example of these improvements in \figref{fig:semantic_completion} where we can observe that our NVC helped ForkNet~\cite{wang2019forknet} classify the table correctly. 
Furthermore, this consequently reduces its gap between ForkNet~\cite{wang2019forknet}, and the state-of-the art methods from ScanComplete~\cite{dai2018scancomplete} and SCFusion~\cite{wu2020scfusion}. 

\begin{figure*}[t]
\centering
\resizebox{1.\linewidth}{!}
{\begin{subfigure}[b]{0.33\linewidth}
\centering
\begin{tikzpicture}
\tikzstyle{every node}=[font=\small]
\begin{axis}[
    title={(a)~MNIST~\cite{lecun1998mnist}},
    xlabel={Number of Anchors},
    ylabel={Absolute Difference},
    xmin=0, xmax=40,
    ymin=0.13, ymax=0.18,
    xtick={0,10,20,30,40},
    ytick={0.13,0.14,0.15,0.16, 0.17, 0.18},
    legend pos=north west,
    legend cell align={left},
    legend style={nodes={scale=0.75, transform shape}},
    ymajorgrids=true,
    grid style=dashed,
    scale=0.55
]

\addplot
    coordinates {
    (1,0.168)(2,0.154)(3,0.143)(4,0.140)(5,0.137)(10,0.136)(15,0.138)(20,0.146)(30,0.158)(40,0.163)
    };
\addlegendentry{NVC \textit{without} $M_{a}$}
\addplot
    coordinates {
    (1,0.167)(2,0.153)(3,0.141)(4,0.136)(5,0.136)(10,0.138)(15,0.139)(20,0.141)(30,0.142)(40,0.142)
    };
\addlegendentry{NVC}
\addplot
    coordinates {
    (1,0.164)(2,0.149)(3,0.141)(4,0.134)(5,0.138)(10,0.133)(15,0.134)(20,0.136)(30,0.136)(40,0.138)
    };
\addlegendentry{NVC-ML}
    
\end{axis}
\end{tikzpicture}
\end{subfigure}

\begin{subfigure}[b]{0.33\linewidth}
\centering
\begin{tikzpicture}
\tikzstyle{every node}=[font=\small]
\begin{axis}[
    title={(b)~NYUv2~\cite{couprie2013indoor}},
    xlabel={Number of Anchors},
    ylabel={Depth Difference},
    xmin=0, xmax=40,
    ymin=0.125, ymax=0.143,
    xtick={0,10,20,30,40},
    ytick={0.125, 0.127, 0.129, 0.131, 0.133, 0.135, 0.137, 0.139, 0.141, 0.143},
    yticklabel style={/pgf/number format/.cd,fixed,precision=3},
    legend pos=north west,
    legend cell align={left},
    legend style={nodes={scale=0.75, transform shape},},
    ymajorgrids=true,
    grid style=dashed,
    scale=0.55
]

\addplot
    coordinates {
    (1, 0.135)(2, 0.135)(3, 0.135)(4, 0.131)(5,0.132)(10,0.131)(15,0.129)(20,0.129)(30,0.134)(40,0.135)
    };
\addplot
    coordinates {
    (1, 0.133)(2, 0.133)(3, 0.133)(4, 0.131)(5,0.130)(10,0.129)(15,0.126)(20,0.127)(30,0.129)(40,0.130)
    };
\addplot
    coordinates {
    (1, 0.133)(2, 0.133)(3, 0.131)(4, 0.130)(5,0.129)(10,0.129)(15,0.126)(20,0.125)(30,0.126)(40,0.129)
    };
\legend{NVC \textit{without} $M_{a}$,NVC,NVC-ML}
    
\end{axis}
\end{tikzpicture}
\end{subfigure}
\begin{subfigure}[b]{0.33\linewidth}
\centering
\begin{tikzpicture}
\tikzstyle{every node}=[font=\small]
\begin{axis}[
    title={(c)~ShapeNet~\cite{shapenet2015}},
    xlabel={Number of Anchors},
    ylabel={IoU},
    xmin=0, xmax=40,
    ymin=68, ymax=84,
    xtick={0,10,20,30,40},
    ytick={68, 70, 72, 74, 76, 78, 80, 82, 84},
    legend pos=south west,
    legend cell align={left},
    legend style={nodes={scale=0.75, transform shape}},
    ymajorgrids=true,
    grid style=dashed,
    scale=0.55
]

\addplot
    coordinates {
    (1,77.5)(2,77.9)(3,78.8)(4,78.0)(5,81.1)(10,79.3)(15,74.9)(20,74.8)(30,75.1)(40,74.0)
    };
\addplot
    coordinates {
    (1,77.4)(2,78.1)(3,78.6)(4,79.8)(5,81.6)(10,82.3)(15,82.4)(20,82.1)(30,81.6)(40,79.3)
    };
\addplot
    coordinates {
    (1,77.9)(2,79.2)(3,79.3)(4,81.0)(5,83.2)(10,82.8)(15,83.0)(20,83.1)(30,82.9)(40,82.2)
    };
\legend{NVC \textit{without} $M_{a}$,NVC,NVC-ML}
    
\end{axis}
\end{tikzpicture}
\end{subfigure}

}
\caption{Plots the performance on the image reconstruction (MNIST)~\cite{lecun1998mnist}, 3D planar reconstruction (NYUv2)~\cite{couprie2013indoor} and 3D object completion (ShapeNet)~\cite{shapenet2015} 
with different amount of nebula anchor.}
\label{fig:linechart_anchor_numbers}
\vspace{15pt}
\end{figure*}
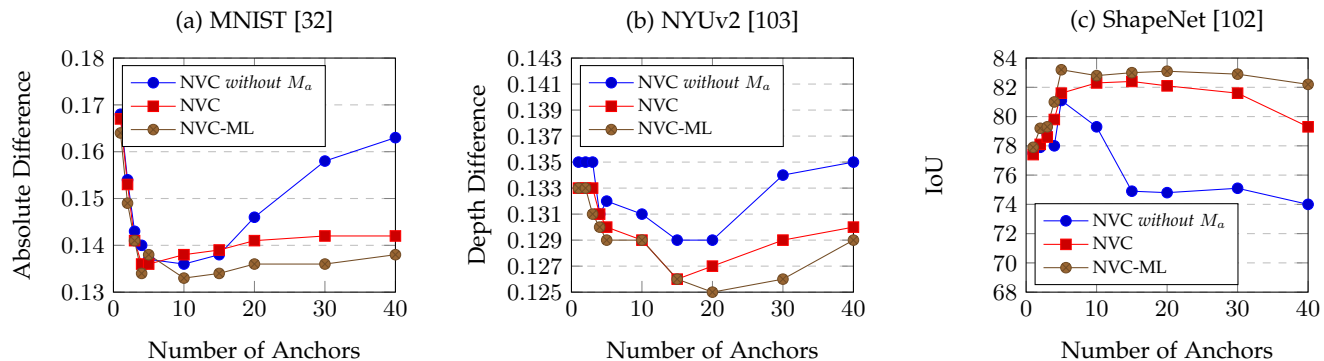

\begin{table*}[!ht]
 \centering
\resizebox{.99\linewidth}{!}
{\begin{tabular}{l|l|ccccccccccc}
	\toprule	
	Dataset  & Method & $\mathcal{G}(\mathcal{E}())$ & 1 & 2 & 3 & 4 & 5 & 10 & 15 & 20 & 30 & 40 \\
	\midrule	
		MNIST~\cite{lecun1998mnist}: rel & NVC \textit{without} $M_{a}$ & 0.169 & 0.168 & 0.154 & 0.143 & 0.140 & 0.137 & \textbf{0.136} & 0.138 & 0.146 & 0.158 & 0.163 \\
		with VO-Hand~\cite{gao2019variational} & NVC & 0.169 & 0.167 & 0.153 & 0.141 & \textbf{0.136} & 0.136 & 0.138 & 0.139 & 0.141 & 0.142 & 0.142 \\ 
		& NVC-ML & 0.169 & 0.164 & 0.149 & 0.141 & 0.134 & 0.138 & \textbf{0.133} & 0.134 & 0.136 & 0.136 & 0.138 \\
	\midrule	
		NYUv2~\cite{couprie2013indoor}: rel & NVC \textit{without} $M_{a}$ & 0.134 & 0.135 & 0.135 & 0.135 & 0.131 & 0.132 & 0.131 & \textbf{0.129} & 0.129 & 0.134 & 0.135   \\ 
        with PlanarRecon~\cite{yu2019single} & NVC & 0.134 & 0.133 & 0.133 & 0.133 & 0.131 & 0.130 & 0.129 & \textbf{0.126} & 0.127 & 0.129 & 0.130 \\
        & NVC-ML & 0.134 & 0.133 & 0.133 & 0.131 & 0.130 & 0.129 & 0.129 & 0.126 & \textbf{0.125} & 0.126 & 0.129 \\
	\midrule	
		ShapeNet~\cite{shapenet2015}: IoU (in \%) & NVC \textit{without} $M_{a}$ & 77.5 & 77.5 & 77.9 & 78.8 & 78.0 & \textbf{81.1} & 79.3 & 74.9 & 74.8 & 75.1 & 74.0 \\ 
        with 3D-RecGAN~\cite{yang2018dense} & NVC & 77.5 & 77.4 & 78.1 & 78.6 & 79.8 & 81.6 & 82.3 & \textbf{82.4} & 82.1 & 81.6 & 79.3 \\
        & NVC-ML & 77.5 & 77.9 & 79.2 & 79.3 & 81.0 & \textbf{83.2} & 82.8 & 83.0 & 83.1 & 82.9 & 82.2 \\
	\bottomrule	
 \end{tabular}}
 \caption{Comparison of different amount of nebula anchors $a$ with and without the self-supervised metric learning, evaluated for the 2D image reconstruction on MNIST~\cite{lecun1998mnist} and the depth estimation on NYUv2~\cite{couprie2013indoor} reported in absolute difference (rel) and 3D object completion on ShapeNet~\cite{shapenet2015} reported in Intersection over Union (IoU).
 \label{tab:ablation_anchors}}
\end{table*}

\section{Understanding the Number of Anchors}
\label{sec:num_anchors}

This section focuses on the hyperparameter of the proposed method which is the number of anchors. 
\figref{fig:linechart_anchor_numbers} and \tabref{tab:ablation_anchors} provide the results with different numbers of nebula anchors.
Since the number of anchors is a crucial hyperparameter, we evaluate how the number of anchors influence the performance of inference model on three datasets --
%
image reconstruction (MNIST)~\cite{lecun1998mnist}, 3D planar reconstruction (ScanNet)~\cite{dai2017scannet} and 3D object completion (ShapeNet)~\cite{shapenet2015}.

When the number of anchors is set to zero, the performance of the inference model is the same as original model. 
Although training with less than 5 anchors starts to improve the performance of the original architectures, the significant improvements are more evident in the range of 5 to 20 anchors depending on the task.

The goal of this section is to conduct an ablation study to investigate the optimum performance in relation to number of anchors.
Ideally, the hyperparameter must have a stable range of optimal values so that we can easily set its value prior to training.
Therefore, this study highlight the influence of $M_{a}$ in \secref{sec:eval_m_a} as well as the influence of the metric learning in \secref{sec:eval_metric} in order to stabilize the optimal range.

%

\subsection{Influence of $M_{a}$}
\label{sec:eval_m_a}

From \secref{sec:nvc}, the mass $M_{a}$ acts as the weight of different clusters centered around the anchors, which depends on how many samples are assign to the anchors. Thus, anchors with a small number of samples back-propagate much smaller gradients. By weighing the loss with the mass, the important clusters are focused during training.

An interesting observation in \figref{fig:linechart_anchor_numbers} is the effects of $M_{a}$ to the range of the optimal number of anchors. We notice that training without $M_{a}$ makes the performance of the inference model sensitive to the number of anchors, reaching the best results only at a certain peak. 
In this case, the mass-based term in $\mathcal{L}_\text{nebula}$ is not used but instead we use the Euclidean distance between the latent samples and their closest anchor to optimize the network.
After investigating \figref{fig:linechart_anchor_numbers}, the plot from ShapeNet demonstrate the worst case scenario where there is only one optimal value for the number of anchors (\ie 5); otherwise, its performance drops significantly. 
With the help of $M_{a}$, the results becomes more stable such that a range of anchor sizes achieve good results instead of one.

\subsection{Influence of the metric learning}
\label{sec:eval_metric}

As described in \secref{sec:supervised}, we can optionally apply metric learning such as Siamese and triplet training on the clusters. 
This becomes more evident in \figref{fig:linechart_anchor_numbers} where NVC with metric learning (NVC-ML) further improves the results. 
Looking more closely at the plots, we notice that the metric learning also stabilizes the range of optimal number of anchors especially for ScanNet~\cite{dai2017scannet} in \figref{fig:linechart_anchor_numbers}.

\begin{table*}[t]
\centering
\resizebox{\linewidth}{!}
{\begin{tabular}{l|l|ccc|ccc}
	\toprule	
	 Dataset & Method & Baseline & \textit{with} \textbf{NVC} & \textit{without} $M_\text{a}$ & \textit{with} \textbf{ML} & \textit{without} $\mathcal{L}_\text{pair}$ & \textit{without} $\mathcal{L}_\text{triplet}$  \\
	\midrule
	 Object Completion on& 3D-AE~\cite{yang2018dense} & 16.6 & 17.1 & 16.7 & \textbf{18.4} & 17.9 & 17.2 \\ 
       Real~\cite{yang2018dense} & 3D-RecGAN~\cite{yang2018dense} & 16.5 & 19.6 & 17.0 & \textbf{20.5} & 18.1 & 17.9 \\ 
       & ForkNet~\cite{wang2019forknet} & 23.8 & 24.9 & 22.1 & \textbf{26.3} & 24.5 & 23.9 \\ 
	\midrule
Semantic Completion on & ForkNet~\cite{wang2019forknet} & 9.3 & 14.1 & 11.5 & \textbf{16.2} & 14.9 & 14.2 \\
  CompleteScanNet~\cite{wu2020scfusion} & ScanComplete~\cite{dai2018scancomplete} & 16.1 & 18.4 & 16.9 & \textbf{19.8} & 17.2 & 17.3 \\
 & SCFusion~\cite{wu2020scfusion} & 18.9 & 20.0 & 19.1 & \textbf{22.6} & 20.7 & 20.5 \\
	\bottomrule
	\end{tabular}
}
\caption{Ablation study on loss functions, comparing the nebula loss against a Euclidean loss (\ie without $M_\text{a}$) and comparing the contribution of the losses metric learning. The results are reported in the average IoU (\%) across different categories for object completion~\cite{yang2018dense} in \tabref{tab:shapenet_iou} and semantic scene completion~\cite{wu2020scfusion} in \tabref{tab:completescannet}.
\label{tab:loss_functions}} 
\end{table*}

\section{Ablation study on the loss function}

Using the experiments from the object completion in \tabref{tab:shapenet_iou} and semantic completion in \tabref{tab:completescannet}, we perform an ablation study on the loss introduced in this paper. The first comparison in \tabref{tab:loss_functions} shows the evaluation when simplify $\mathcal{L}_\text{nebula}$ to a Euclidean distance (labelled as without $M_\text{a}$). This change based on ForkNet~\cite{wang2019forknet} produces a noticeable reduction in performance by 2.8\% in terms of IoU for object completion and 2.6\% for semantic scene completion.

The second comparison shows the advantage of having  the components $\mathcal{L}_\text{pair}$ and $\mathcal{L}_\text{triplet}$ in metric learning. 
\tabref{tab:loss_functions} validates that without either of the loss, the IoU is reduced by up to 2.4\% for object completion with ForkNet~\cite{wang2019forknet} and 2.5\% with ScanComplete. Note that the results indicate that training without $\mathcal{L}_\text{triplet}$ introduce larger performance deduction compared to training without $\mathcal{L}_\text{pair}$.


\section{Conclusion}

Focused on self-supervised latent space optimization, we present a novel nebula variational coder based on two main contributions: 
(i) forming clusters in the latent space through the additional variables, called nebula anchors, trained in an unsupervised way; and, 
(ii) by labeling features with the assigned anchors, employing metric learning to further separate the clusters in a self-supervised way. 
The proposed approach showed, on one side, the performance gains when tested on supervised and unsupervised tasks and, on the other, good generalization capabilities to deal with different network architectures and data dimensionality.

\bibliographystyle{IEEEtran}
\bibliography{vcclv}



%

\begin{IEEEbiography}[{\includegraphics[width=1in,height=1.25in,clip,keepaspectratio]{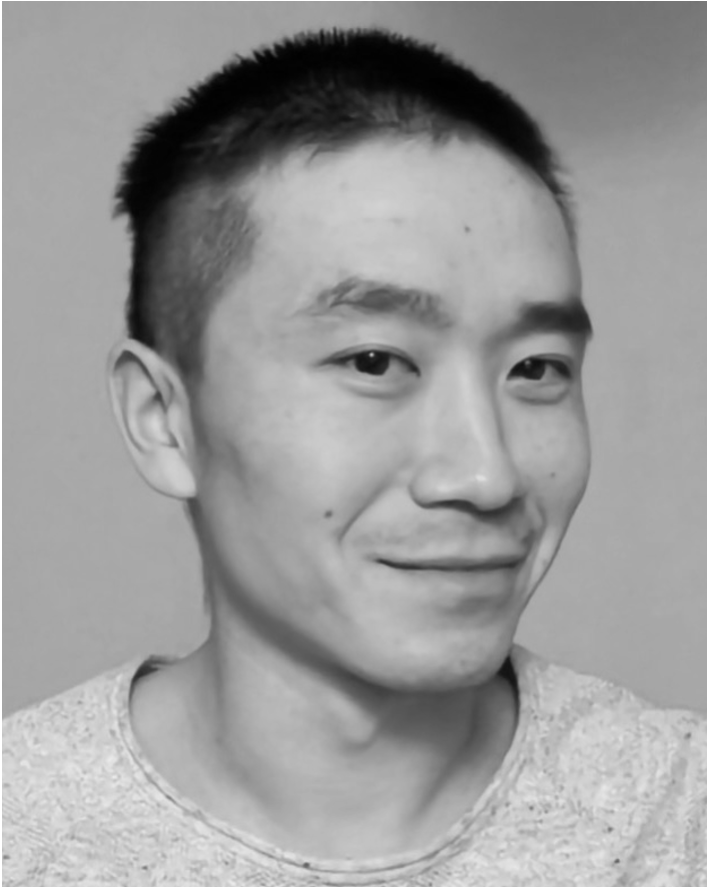}}]{Yida Wang}
had received his Ph.D. from Technische Universit\"at M\"unchen (TUM) focusing on 3D understanding and reconstruction based on deep architectures and statistical approaches. His research activities have been published on prestigious conferences such as CVPR, ICCV, ECCV, 3DV, IROS, ICRA and journals such as IJCV, TIP and RA-L. 
\end{IEEEbiography}

\begin{IEEEbiography}[{\includegraphics[width=1in,height=1.25in,clip,keepaspectratio]{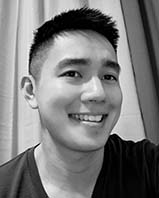}}]{David Joseph Tan}
is a Research Scientist at Google. He got his Ph.D.~from the Technische Universit\"{a}t M\"{u}nchen (TUM). In 2017, the culmination of his Ph.D.~research on the ultra-fast 6D pose estimation garnered him the Best Demo Award at ISMAR, EXIST-Gr\"{u}nderstipendium from the German government and the Promotionspreise from TUM.
Expanding his research in computer vision and machine learning, he continuously publishes in top-tier conferences and journals.
\end{IEEEbiography}

\vfill

\begin{IEEEbiography}[{\includegraphics[width=1in,height=1.25in,clip,keepaspectratio]{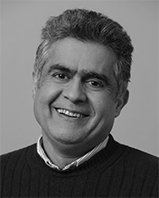}}]{Nassir Navab}
is a full professor and director of the Laboratories for Computer Aided Medical Procedures at Technical University of Munich (TUM) and Johns Hopkins University with secondary faculty appointments at both Medical Schools. He is also the director of Medical Augmented Reality summer school series at Balgrist Hospital in Zurich. He received the prestigious Siemens Inventor of the Year Award in 2001. He also received the SMIT Technology Award in 2010 and IEEE ISMAR 10 Years Lasting Impact Award in 2015. He had received his PhD from INRIA and University of Paris XI in France and enjoyed two years postdoctoral fellowship at MIT Media Laboratory before joining SCR in 1994. He is Fellow of the MICCAI Society and asked on its board of directors from 2007 to 2012 and from 2014 to 2017. He served as General Chair for MICCAI 2015, ISMAR 2001, 2005 and 2014. He is a funding board member of IPCAI 2010-2021 and Area Chair for ICCV 2022 and ECCV 2020. He is on the editorial board of many international journals including IEEE TMI and MedIA. As of March 2022, his papers have received over 50000 citations and enjoy an h-index of 100. 
\end{IEEEbiography}

\begin{IEEEbiography}[{\includegraphics[width=1in,height=1.25in,clip,keepaspectratio]{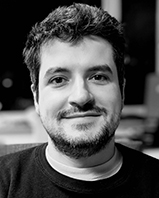}}]{Federico Tombari}
is a Research Scientist and manager at Google where he leads an applied research team in computer vision and machine learning. He is also a Lecturer (PrivatDozent) at the Technical University of Munich (TUM). He has 200+ peer-reviewed publications in the field of 3D computer vision and machine learning and their applications to robotics, autonomous driving, healthcare and augmented reality. He got his PhD in 2009 from the University of Bologna, where he was Assistant Professor from 2013 to 2016, and his Habilitation from TUM in 2018. In 2018-19, he was co-founder and managing director of a Munich-based startup on 3D perception for AR and robotics. He regularly serves as Chair and Associate Editor for international conferences and journals (RA-L, ECCV18, IROS20, ICRA20, IROS21, 3DV19, 3DV20, 3DV21 among others). He was the recipient of two Google Faculty Research Awards, one Amazon Research Award, two CVPR Outstanding Reviewer Awards. He has been a research partner of private and academic institutions including Google, Toyota, BMW, Audi, Amazon, Univ. Stanford, ETH and JHU.
\end{IEEEbiography}

\vfill







\end{document}